\documentclass[10pt,twocolumn,letterpaper]{article}

\usepackage{cvpr}              
\usepackage{graphicx}
\usepackage{amsmath}
\usepackage{amssymb}
\usepackage{booktabs}
\usepackage{arydshln}
\usepackage[accsupp]{axessibility} 
\usepackage{multirow}
\usepackage{animate}
\usepackage[dvipsnames]{xcolor}
\usepackage{pifont}
\newcommand{\cmark}{\ding{51}}%
\newcommand{\xmark}{\ding{55}}%
\newcommand{\FilledCircle}{\ding{108}}
\usepackage[misc]{ifsym}
\usepackage{insertpdf}

\usepackage[pagebackref,breaklinks,colorlinks]{hyperref}

\usepackage[capitalize]{cleveref}
\crefname{section}{Sec.}{Secs.}
\Crefname{section}{Section}{Sections}
\Crefname{table}{Table}{Tables}
\crefname{table}{Tab.}{Tabs.}

\def\cvprPaperID{674} 
\def\confName{CVPR}
\def\confYear{2022}

\begin{document}
\title{No Pain, Big Gain: Classify Dynamic Point Cloud Sequences with Static Models\\by Fitting Feature-level Space-time Surfaces}
\author{Jia-Xing Zhong, Kaichen Zhou, Qingyong Hu${^{\footnotesize{\textrm{\Letter}}}}$, Bing Wang, Niki Trigoni, Andrew Markham\\
Deparment of Computer Science, University of Oxford\\
{\tt\small \{jiaxing.zhong, rui.zhou, qingyong.hu, bing.wang, niki.trigoni, andrew.markham\}@cs.ox.ac.uk}
}
\maketitle
\begin{abstract}
Scene flow is a powerful tool for capturing the motion field of 3D point clouds. However, it is difficult to directly apply flow-based models to dynamic point cloud classification since the unstructured points make it hard or even impossible to efficiently and effectively trace point-wise correspondences. To capture 3D motions without explicitly tracking correspondences, we propose a \textbf{kine}matics-inspired neural \textbf{net}work (Kinet) by generalizing the kinematic concept of ST-surfaces to the feature space. By unrolling the normal solver of ST-surfaces in the feature space, Kinet implicitly encodes feature-level dynamics and \underline{gains} advantages from the use of mature backbones for static point cloud processing. With only minor changes in network structures and low computing overhead, it is \underline{painless} to jointly train and deploy our framework with a given static model. Experiments on {NvGesture}, {SHREC'17}, {MSRAction-3D}, and {NTU-RGBD} demonstrate its efficacy in performance, efficiency in both the number of parameters and computational complexity, as well as its versatility to various static backbones. Noticeably, Kinet achieves the accuracy of 93.27\% on \textit{MSRAction-3D} with only 3.20M parameters and 10.35G FLOPS. The code is available at \href{https://github.com/jx-zhong-for-academic-purpose/Kinet}{https://github.com/jx-zhong-for-academic-purpose/Kinet}.
\end{abstract}
\section{Introduction}
Due to continued miniaturization and mass production, 3D sensors are becoming less esoteric and increasingly prevalent in geometric perception tasks. These sensors typically represent scene geometry through a point cloud, which is an unordered and irregular data structure consisting of distinct spatial 3D coordinates. As a fundamental problem in point cloud understanding, classification of static scenes~\cite{scannet, hu2019randla, thomas2019kpconv} or objects~\cite{qi2017pointnet, qi2016volumetric, chang2015shapenet} has witnessed rapid advances over the past few years. Whilst impressive, these techniques do not directly account for the fact that the real 3D world is also changing, through egocentric and/or allocentric motion. To better understand our time-varying world, a handful of recent works~\cite{liu2019meteornet,fan2019pointrnn,fan2021pstnet,fan21p4transformer,efficient_pointlstm,min2019flickernet,wei2021spatial} have been applied to dynamic point cloud classification, a task in which the model is required to output a video-level category for a given sequence of 3D point clouds. 

\begin{figure}[t]
\centering
\captionsetup[subfigure]{justification=centering, skip=5pt}
\begin{subfigure}{0.48\textwidth}
\centering   
\includegraphics[width=.75\linewidth]{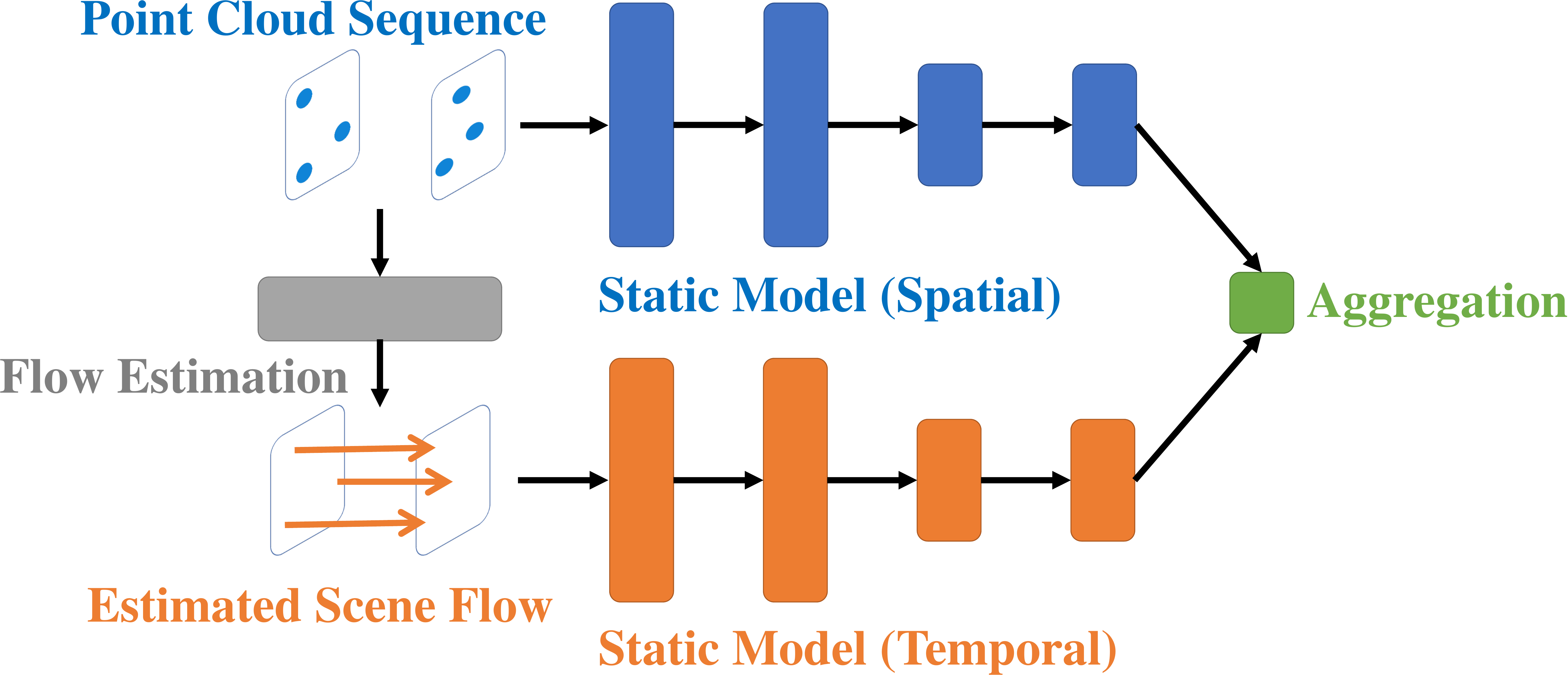}
\caption{Vanilla two-stream framework based on physical scene flow.}\label{fig:raw_framework}
\end{subfigure}

\vspace{0.3cm}
\begin{subfigure}{0.48\textwidth}
\centering   
\includegraphics[width=.75\linewidth]{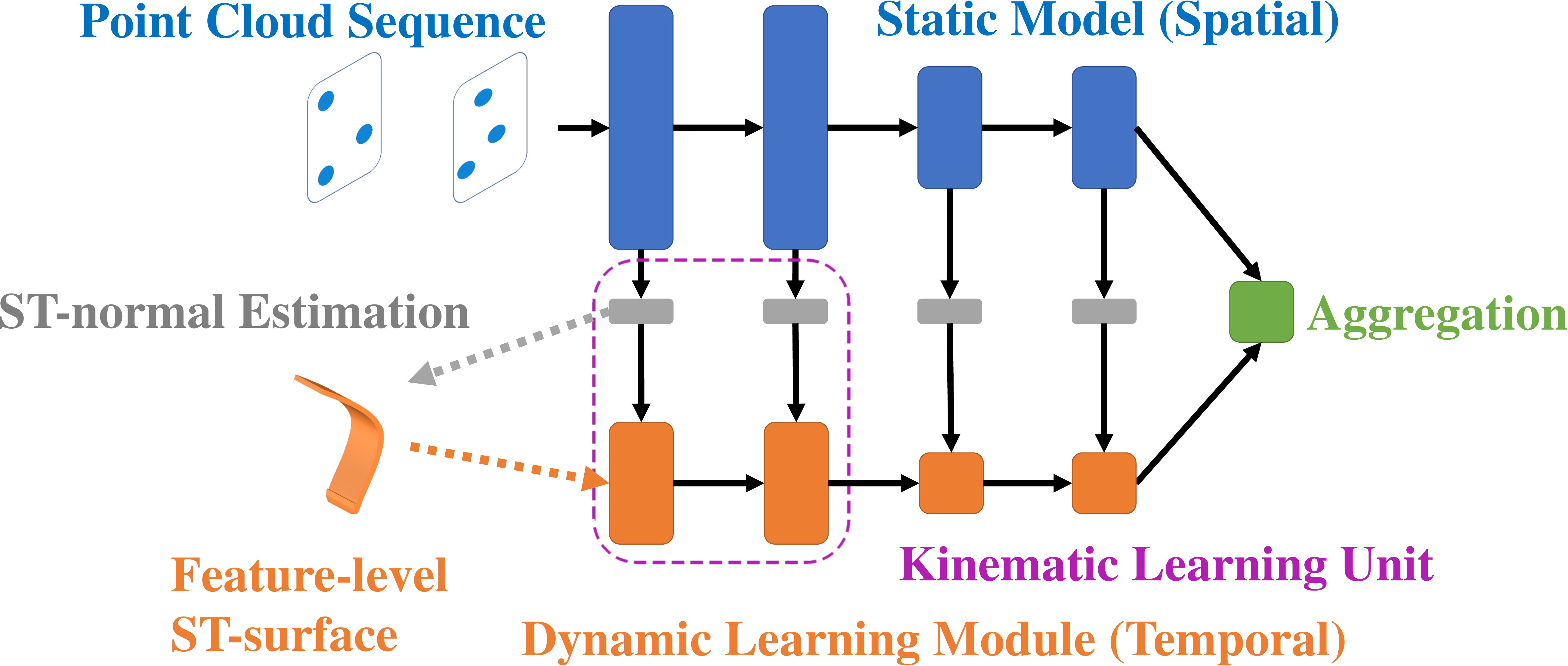}
\caption{Kinematic two-stream framework guided by feature-level ST-surfaces.}\label{fig:our_framework}
\end{subfigure}
\caption{\textit{Comparison between the flow-based framework and ours.} With neither explicit point-wise correspondence estimation nor the stand-alone temporal branch, our framework is lightweight and efficient.}
\label{fig:comp_framworks} 
\end{figure}

As a natural extension of 2D optical flow, 3D scene flow captures the motion field of point clouds. Based on optical flow, two-stream networks~\cite{NIPS2014_twostream,feichtenhofer2016convolutional_2fusion,zhang2020v4d,Carreira_2017_i3d,TSN2016ECCV} have already proven to be successful in image-based video classification. Hence, it should be a natural choice to classify dynamic point clouds with the help of scene flow. However, to the best of our knowledge, scene flow has not been utilized in point cloud sequences despite the prevalence of mature scene flow estimators~\cite{justgo,liu:2019:flownet3d,HPLFlowNet,gojcic2021weakly,puy20flot,behl2019pointflownet,tishchenko2020self,wu2020pointpwc}.

\textit{What then hampers us from applying scene flow to dynamic point cloud classification?} Although scene flow is a powerful tool, it is difficult to estimate it efficiently and effectively from sequential point clouds - the computation of 3D scene flow inevitability has higher time expenditure, larger memory consumption, and lower accuracy than that of 2D optical flow. These challenges are mainly caused by the irregular and unordered nature of dynamic point clouds. This unstructured nature makes it difficult to track the point-wise correspondences of the moving point sets across different frames.

\textit{Why not extract dynamic information without explicitly finding the point-wise correspondences?} If this were possible, researchers could gain advantages from the decoupled motion representations but not suffer from the painful computational process of scene flow. Similar to the \underline{gains} seen in two-stream networks in image-based models, we would be able to \textit{preserve the benefit of mature static solutions} in inference and training, such as well-benchmarked network architectures, transferable pre-trained weights, and ready-to-use source code. At the same time, the \underline{pain} of scene flow estimation would be significantly relieved, only with \textit{minor network modification and low computational overhead}.

For this purpose, we get inspiration from kinematics and propose a neural network (\textbf{Kinet}) to bypass direct scene flow estimation by generalizing the kinematic concept of space-time surfaces~\cite{pottmann2001computational} (ST-surfaces) from the physical domain of point clouds to the feature space. In this way, normal vectors \wrt these ST-surfaces (ST-normals) establish the representation field of dynamic information as shown in Figure~\ref{fig:our_framework}. Thus, \textit{motions are implicitly represented by feature-level ST-surfaces without explicitly computing point-wise correspondences}. Inspired by iterative normal refinement~\cite{mitra2007dynamic}, we unroll the solver for ST-normals and make it jointly trainable alongside the static model in an end-to-end manner. Inheriting intermediate features from static network layers, Kinet is lightweight in parameters and efficient in computational complexity, compared with the vanilla flow-based framework depicted in Figure~\ref{fig:raw_framework} which requires extra scene flow estimation and the independent temporal branch.

Experiments are conducted on four datasets (\textit{NvGesture} \cite{r3dcnn}, \textit{SHREC'17} \cite{de2017shrec}, \textit{MSRAction-3D} \cite{MSRAction-3D} and \textit{NTU-RGBD}~\cite{shahroudy2016ntu}) for two tasks (gesture recognition and action classification) with three typical static backbones (MLP-based \textit{PointNet++} \cite{qi2017pointnet++}, graph-based \textit{DGCNN} \cite{dgcnn} and convolution-based \textit{SpiderCNN}~\cite{xu2018spidercnn}). Noticeably, 1) in gesture recognition, our framework outperforms humans for the first time with the accuracy of $89.1\%$ on \textit{NvGesture}; 2) in action classification, it achieves a new record of $93.27\%$ on 24-frame \textit{MSRAction-3D} with only $3.20M$ parameters and $10.35G$ FLOPS.

In summary, our main contribution is as follows:
\begin{itemize}
\setlength{\itemsep}{0pt}
\setlength{\parsep}{0pt}
\setlength{\parskip}{0pt}
\item By introducing Kinet, we decouple temporal information from spatial features, thereby easily extending static backbones to dynamic recognition and entirely preserving the merits of these mature backbones.
\item Without the pain of tracking point-wise correspondences, we encode point cloud dynamics by unrolling the ST-normal solver in the feature space. This method is jointly trainable alongside the static model, with minor structural changes and low computing overhead.
\item Extensive experiments on various datasets, tasks, and static backbones show its efficacy in performance, efficiency in parameters and computational complexity, as well as versatility to different static backbones. The code is available at \href{https://github.com/jx-zhong-for-academic-purpose/Kinet}{https://github.com/jx-zhong-for-academic-purpose/Kinet}.
\end{itemize}
\section{Related Work}
\noindent\textbf{Deep Learning on Static Point Clouds} Recently, deep learning on 3D point clouds has attracted increased attention~\cite{guo2020deep}, with substantial progress achieved in several fields including shape classification~\cite{qi2017pointnet, li2018pointcnn, chang2015shapenet, mo2019partnet}, object detection~\cite{caesar2020nuscenes, lang2019pointpillars, qi2018frustum, shi2019pointrcnn} and scene segmentation~\cite{scannet, behley2019semantickitti, SensatUrban, hu2019randla, 3D-bonet}. This can be mainly attributed to the availability of various high-quality datasets~\cite{behley2019semantickitti, hu2022sensaturban, scannet} and sophisticated neural architectures~\cite{qi2017pointnet, qi2017pointnet++, li2018pointcnn, hu2021sqn}. From the perspective of scene representations, existing works can be roughly divided into 1) Voxel-based 
methods~\cite{maturana2015voxnet, zhou2018voxelnet, riegler2017octnet, 4dMinkpwski, sparse}, 2) Projection-based methods~\cite{chen2017multi, mvcnn}, 3) Point-based methods~\cite{qi2017pointnet, qi2017pointnet++, li2018pointcnn, dgcnn, xu2018spidercnn, hu2021learning}, and 4) Hybrid methods~\cite{Point-Voxel_CNN, dai20183dmv, qi2016volumetric}. Based on the well-developed static classification models, we attempt to apply them to dynamic point cloud recognition with minor structural surgery and low computational overhead. 

\noindent\textbf{Deep Learning on Dynamic Point Clouds} A handful of recent works have explored dynamic problems on point clouds, such as recognition~\cite{liu2019meteornet,fan2021pstnet,fan21p4transformer,efficient_pointlstm,wei2021spatial}, detection~\cite{huangrui,yin2020lidar,Qi_2021_CVPR}, tracking~\cite{giancola2019leveraging,qi2020p2b}, prediction~\cite{rempe2020caspr,RempeDynamics2020,Weng2020_SPF2,niemeyer2019occupancy,jiang2021learning} and scene flow estimation~\cite{justgo,liu:2019:flownet3d,HPLFlowNet,gojcic2021weakly,puy20flot,behl2019pointflownet,tishchenko2020self,wu2020pointpwc}. Existing works on sequence classification are based on convolutional~\cite{liu2019meteornet,liu2019meteornet,min2019flickernet}, recurrent~\cite{efficient_pointlstm}, self-attentional~\cite{fan21p4transformer,wei2021spatial}, or multi-stream neural networks~\cite{wang20203dv}. As a convolutional framework, MeteorNet~\cite{liu2019meteornet} modeled point cloud dynamics via spatio-temporal neighbor aggregations~\cite{liu2019meteornet}. Likewise, PSTNet~\cite{fan2021pstnet} applied point spatio-temporal convolutions to capture information along the time dimension and the space domain. Derived from recurrent networks, PointLSTM~\cite{efficient_pointlstm} updated the hidden states with the combination of past and current features. Fan \etal~\cite{fan21p4transformer} and Wei \etal~\cite{wei2021spatial} adopted self-attentional structures along with the popularity of video transformers~\cite{girdhar2019video}. By extracting the offline dynamic voxel, 3DV~\cite{wang20203dv} encoded motions and appearances through multiple streams. Our Kinet shares the same idea of decoupling spatial and temporal information as 3DV, but the presented framework requires neither the offline motion extraction nor the extra stand-alone temporal stream.
\begin{figure*}[htbp]
\centering
\begin{minipage}{0.35\textwidth}
\centering
\includegraphics[width=0.7\linewidth]{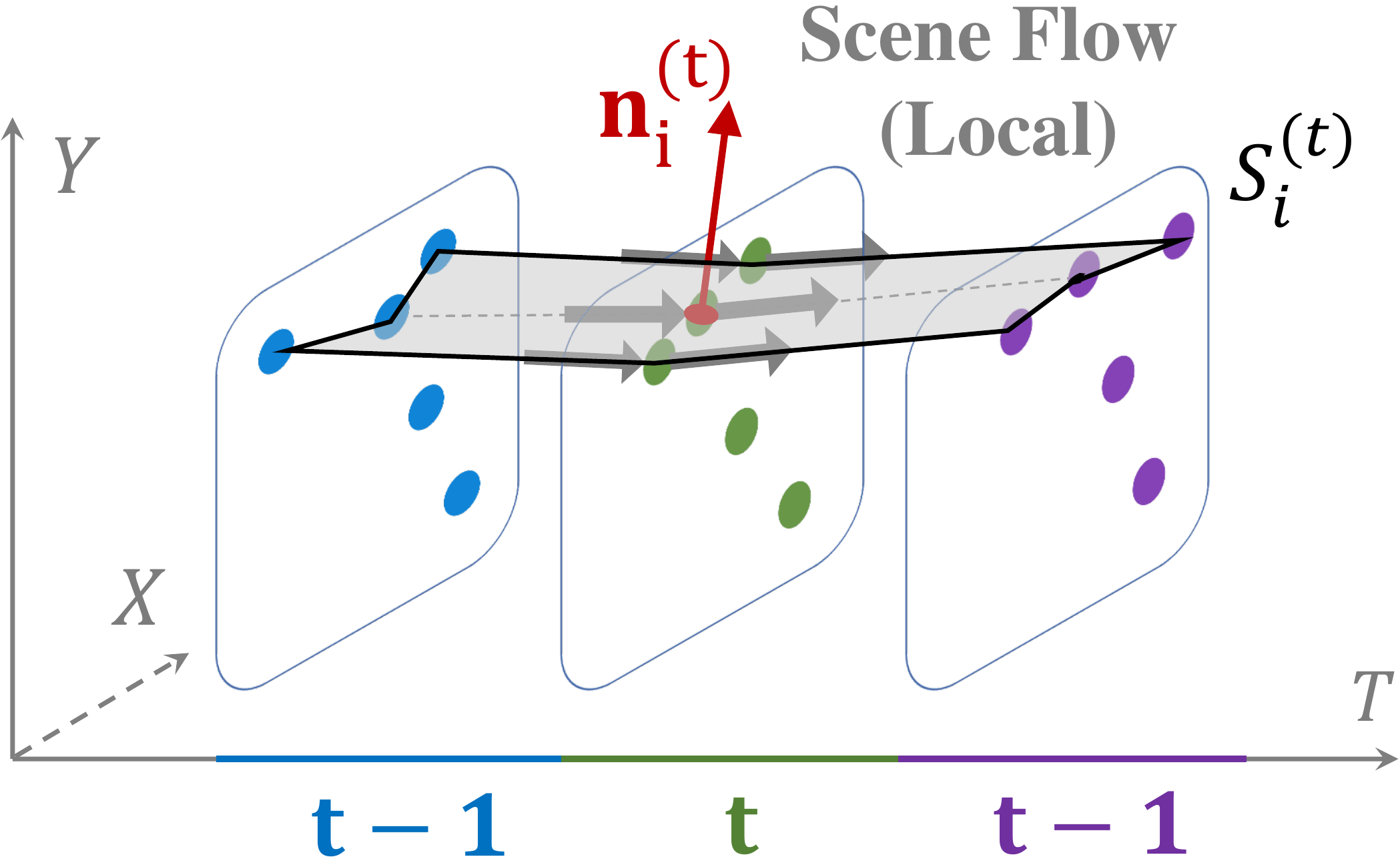}
    \caption{\textit{2D local ST-surface and its normal.} Scene flow (\textcolor{darkgray}{darkgray arrows}) lies on ST-surface $S_i^{(t)}$ and it is orthogonal to the normal $\mathbf{n}_i^{(t)}$.\newline}\label{fig:ST_surface}
\centering
\includegraphics[width=.88\textwidth]{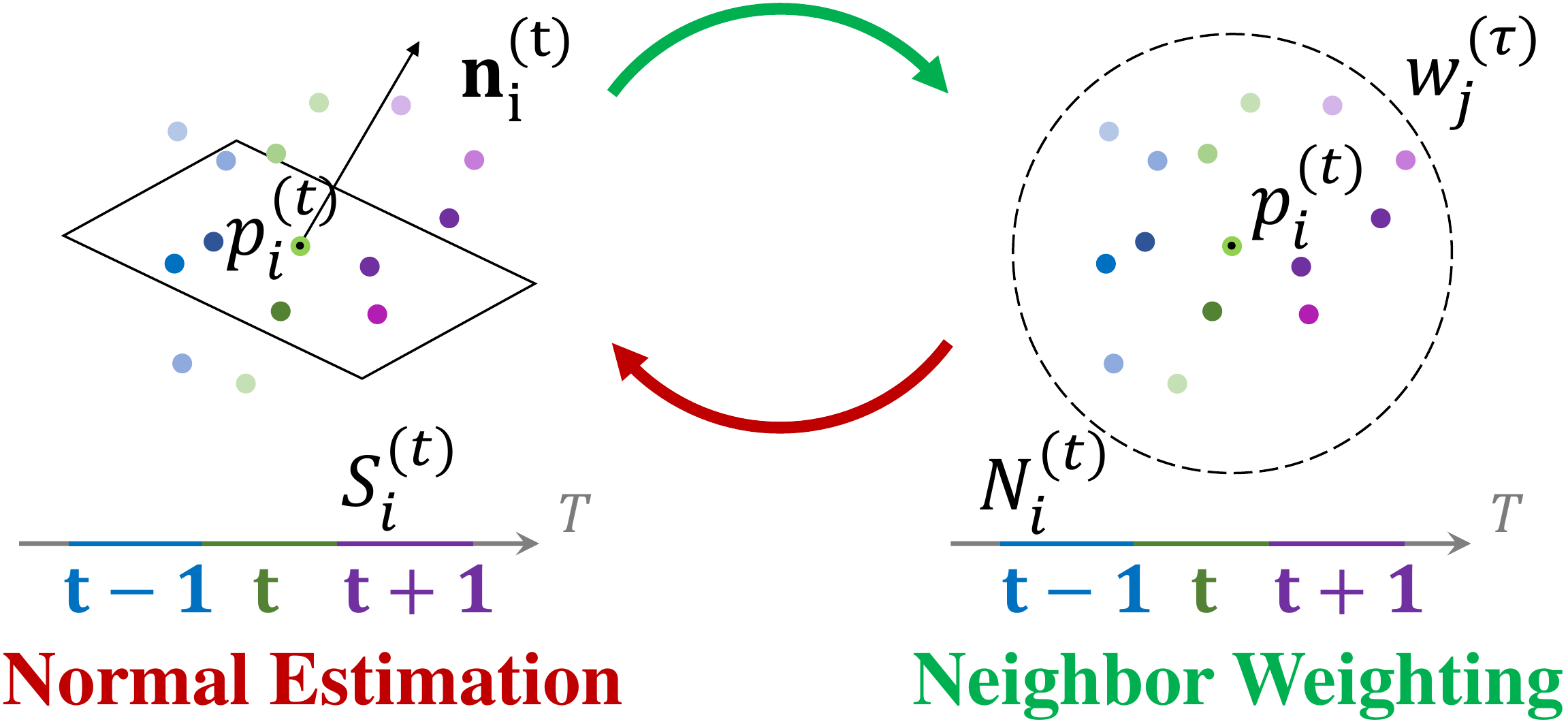}
\caption{\textit{Solver of iterative normal refinement.} The darkness of colors represents weights of neighbor points. The ST-surface is estimated upon the point-wise weights, and vice versa.}\label{fig:normal_refinement}
\end{minipage}~~~~~~~~
\begin{minipage}{0.65\textwidth}
\centering
\includegraphics[width=.98\textwidth]{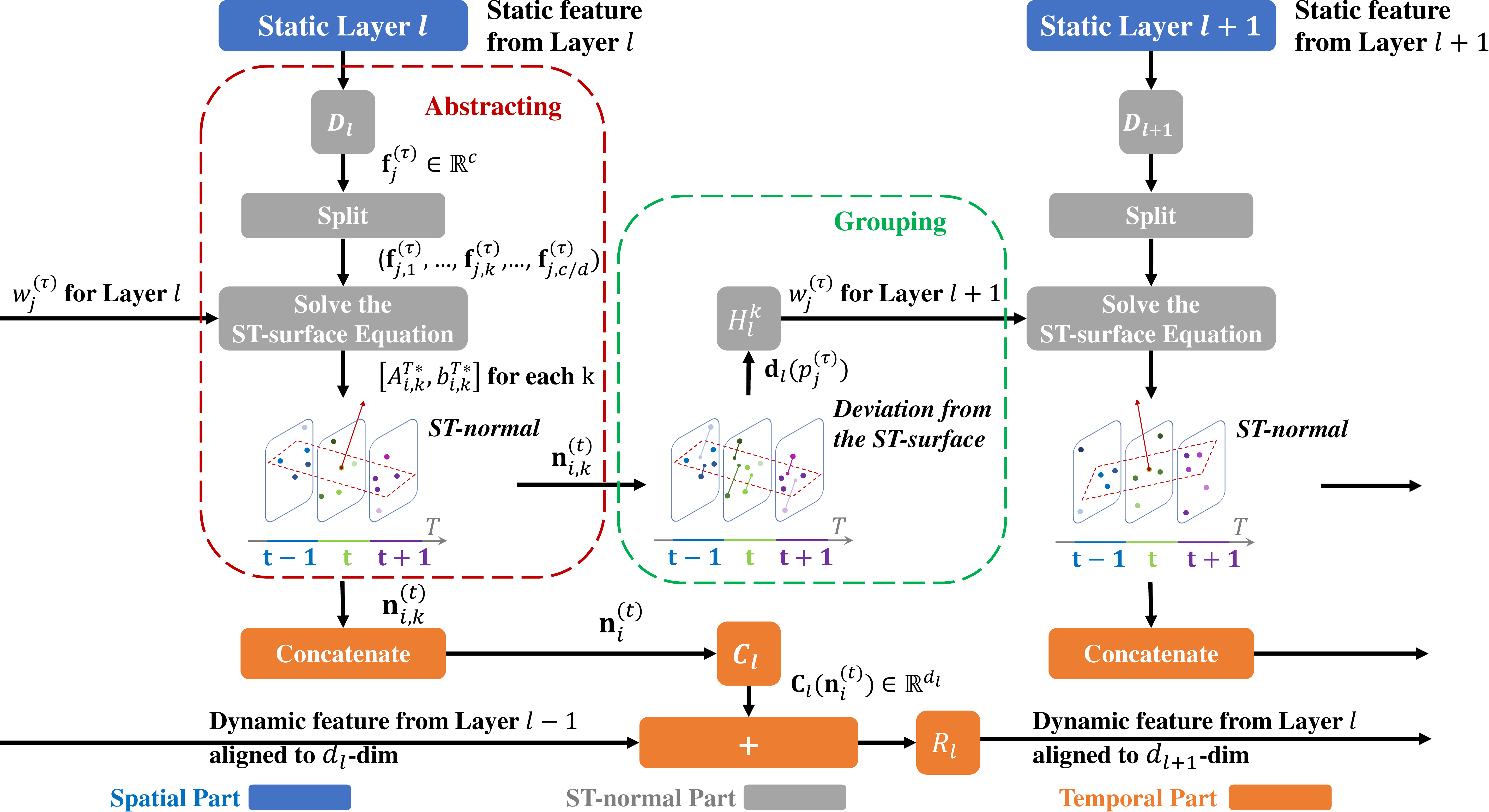}
\caption{\textit{Kinematic learning unit.} As shown in the \textcolor{violet}{violet dotted frame} of Figure~\ref{fig:our_framework}, a stack of these units comprises the temporal branch of our framework. The alternate abstracting (\textcolor{Mahogany}{red dotted frame}) and grouping (\textcolor{ForestGreen}{green dotted frame}) unroll the ST-normal solver (Figure~\ref{fig:normal_refinement}) in the feature space. Only the \textbf{squares} are parametric operations whereas the \textbf{rectangles} introduce no parameters - kinematic learning just requires a small number of learnable parameters.}\label{fig:framework}
\end{minipage}
\end{figure*}

\noindent\textbf{Flow-guided Classifiers for Image-based Videos} The presented work obtains inspiration from the similar idea of encoding optical flow information into deep representations for image-based video classification~\cite{sun2018off,jiang2019stm,PAN,spatial_shift-lee2018motion,lin2019tsm,MotionSqueeze-4_21}. By subtracting feature maps along the temporal axis, OFF~\cite{sun2018off}, STM~\cite{jiang2019stm} and PAN~\cite{PAN} robustly imitated optical-flow calculations. Similarly, Piergiovanni \& Ryoo~\cite{ajpiergiovanni2019representation} and Fan \etal~\cite{TVNet} mimicked TV-L1 optical flow iterations~\cite{tvl1} inside network layers. Heeseung \etal~\cite{MotionSqueeze-4_21} introduced correlations to continuous feature maps. For the purpose of acceleration, temporal shift modules~\cite{lin2019tsm} or spatial shift filters~\cite{spatial_shift-lee2018motion} were utilized to model multi-frame interactions. The above methods rely on feature maps retaining spatial correspondences of regular pixels in images, whereas features of irregular point clouds usually cannot manifest point-wise correspondences across frames. Consequently, all of the aforementioned feature-level operations (subtraction, shift, correlation, \etc) are ineffective for point cloud models. To achieve the similar goal of encoding dynamics from static features, we propose a distinct approach from them. 
\vspace{-0.14cm}\section{Methodology}
Denote an input point cloud sequence with $T$ frames as $P=(P_1,P_2,...,P_{T-1},P_T)$. The $t^{th}$ frame $P_t=\{{p}_i^{(t)}|i=1,2,...,m_t-1,m_t\}$ is a set of $m_t$ points, in which the position of the $i^{th}$ point ${p}_i^{(t)}$ is specified by its spatial coordinates $\mathbf{x}_i^{(t)} =(x_i^{(t)},y_i^{(t)},z_i^{(t)})\in \mathbb{R}^3$. The goal of dynamic point cloud classification is to output the sequence-level category label $y$ for a particular input $P$.
\subsection{Background: Kinematic ST-Surface}
Our method extends the kinematic concept of ST-surfaces~\cite{pottmann2001computational} and adopts the solver of iterative normal refinement~\cite{mitra2007dynamic} from the 3D physical world to deep representation learning.

Intuitively, the local \textbf{ST-surface} $S_i^{(t)} \subset \mathbb{R}^4$ centered at the point $p_i^{(t)}$ is \textit{a surface that fits as many as possible space-time neighbor points} of $p_i^{(t)}$. Figure~\ref{fig:ST_surface} illustrates ST-surface of 2D dynamic point clouds within 3 frames, which can be easily generalized to the 3D case and more frames. According to spatial kinematics~\cite{pottmann2001computational}, instantaneous velocity vectors always lie on the ST-surface. Equivalently, for a point cloud sequence as depicted in Figure~\ref{fig:ST_surface}, Mitra \etal~\cite{mitra2007dynamic} point out that \textit{local scene flow lies on the same ST-surface} due to the vicinal (\ie, local neighbourhood) consistency of movements. As a result, the ST-normal $\mathbf{n}_i^{(t)}$ is orthogonal to local scene flow\footnote{Strictly speaking, the ST-normal $\mathbf{n}_i^{(t)}$ is orthogonal to the local tangent plane of scene flow.} and the field of those normals describes the motions of sequential point clouds. 

Mathematically, the space-time neighbors of $p_i^{(t)}$ are a point set: $N_{\Delta r}^{\Delta t}(p_i^{(t)})=\{{p}_j^{(\tau)}|\; |t- \tau| \leq \Delta t, ||\mathbf{x}_i^{(t)}-\mathbf{x}_j^{(\tau)}||\leq\Delta r\}$, hereinafter referred to as $N_i^{(t)}$ for simplicity. $S_i^{(t)}$ is specified by its tangent plane with the surface equation $A\mathbf{x}+b=t$, of which the coefficient $A\in\mathbb{R}^{1\times 3}$ and $b$ satisfy:
\begin{equation}\label{eq:tangent_plane}
A\mathbf{x}_j^{(\tau)}+b=\tau,~~ \forall{{p}_j^{(\tau)}\in N_i^{(t)}}\,.
\end{equation}

In practice, Equation~(\ref{eq:tangent_plane}) may be an over-determined linear system since the local neighbor area may be too large to reflect the instantaneous velocity. In this case, space-time neighbors $N_i^{(t)}$ cannot be completely represented by coefficients of the tangent plane and there exists no exact solution. Therefore, a least-squared approximation is introduced to seek the optimal coefficients $A^*$ and $b^*$: 
\begin{equation}
  A^*,b^*=\mathop{arg\,min}_{A,b}\sum_{{p}_j^{(\tau)}\in N_i^{(t)}}||A\mathbf{x}_j^{(\tau)}+b-\tau||^2\,.
\end{equation}
To alleviate the influence of noisy point clouds, a commonly-used objective is to obtain the coefficients based on weighted neighbor points:
\begin{equation}\label{eq:w_normal}
  A^*,b^*=\mathop{arg\,min}_{A,b}\sum_{{p}_j^{(\tau)}\in N_i^{(t)}}w_j^{(\tau)}||A\mathbf{x}_j^{(\tau)}+b-\tau||^2\,,
\end{equation}
where $w_j^{(\tau)}$ is the point-wise weight of these neighbors. As a solver of Equation~(\ref{eq:w_normal}), \textbf{iterative normal refinement}~\cite{mitra2007dynamic} robustly encodes dynamics via the normal field of ST-surfaces. This was earliest used in Dynamic Geometry Registration~\cite{mitra2007dynamic}, a traditional method to register large-scale moving and deforming point clouds. As shown in Figure~\ref{fig:normal_refinement}, the basic idea is to alternately re-compute the ST-surface $S_i^{(t)}$ and its normal $\mathbf{n}_i^{(t)}$ based on the neighbors' weights $w_j^{(\tau)}$ and re-weight the space-time neighbors $N_i^{(t)}$ based on the estimated ST-surface $S_i^{(t)}$ until convergence.
\subsection{Kinematic Representation Learning}
\subsubsection{Framework}
The vanilla flow-based framework (Figure~\ref{fig:raw_framework}) explicitly extracts scene flow (or dynamic voxels as in the case of 3DV~\cite{wang20203dv}), while Kinet implicitly encodes motions with feature-level ST-surfaces. As shown in Figure~\ref{fig:our_framework}, Kinet contains three parts: 1) a spatial branch (marked in \textcolor{NavyBlue}{blue}) identical to common static models, 2) a temporal model comprised of stacked kinematic learning units (the \textcolor{violet}{violet dotted frame}), followed by 3) the final aggregation (marked in \textcolor{Green}{green}) of spatial and temporal results.
\subsubsection{Kinematic Learning Unit}
Typically, a learning unit for point clouds has two crucial operations, \ie, \textbf{grouping} and \textbf{abstracting}. The former selects the neighbors around centroids (\eg ball query in PointNet++), while the latter encodes the local feature from these neighbors (\eg PointNet layers in PointNet++). Figure~\ref{fig:normal_refinement} demonstrates that iterative normal refinement alternates between two similar operations: \textbf{neighbor weighting} corresponding to grouping and \textbf{normal estimation} analogous to abstracting. The original integrative normal refinement works in the non-differentiable physical space of three-dimensional point sets - we unroll its solver in a fully \textit{differentiable} fashion and generalize it to the \textit{high-dimensional} feature space for joint optimization in neural networks as depicted in Figure~\ref{fig:framework}.

Assume that we obtain a series of features $F_l(P)=F_l(P_1),F_l(P_2),...,F_l(P_{T-1}),F_l(P_T)$ from the $l^{th}$ layer (marked in \textcolor{NavyBlue}{blue} in Figure~\ref{fig:framework}) of a certain static model $F$. Based on sequential static features $F_l(P)$, our learning unit aims to obtain dynamic representations by fitting feature-level ST-surfaces. 
\paragraph{Abstracting with Normal Estimation (\textcolor{Mahogany}{Red Dotted Frame} in Figure~\ref{fig:framework})}To decrease the computational complexity, we first utilize a 1$\times$1 convolution $D_l$ to reduce its dimension to $c$, where $c$ is proportional to the dimension of a static feature. For a given point $p_i^{(t)}$, the corresponding $c-$dimensional feature vector is denoted as $\mathbf{f}_i^{(t)}\in \mathbb{R}^c$. Similar to the physical space, the tangent hyper-plane of ST-surfaces in the feature space is specified by its surface equation $A\mathbf f+b=t$, where the coefficients $A$ and $b$ satisfy: 
\begin{equation}\label{eq:feat_surface}
A\mathbf{f}_j^{(\tau)}+b=\tau,~~ \forall{{p}_j^{(\tau)}\in N_i^{(t)}}\,.
\end{equation}
Likewise, \textit{time-varying changes of those static features lie on the corresponding ST-(hyper)surface in the representation space}. The vector field of normals w.r.t. such feature-level ST-surfaces orthogonally describes the dynamic information based on static representations.

The equation of ST-surfaces in the 3D physical space is usually over-determined ($|N_i^{(t)}|>3$), whereas it is not the case in the $c-$dimensional feature space ($|N_i^{(t)}|<c$). To ensure exact coefficient solutions in Equation~\ref{eq:feat_surface}, we split the point-wise feature $\mathbf{f}_j^{(\tau)}$ into several $d-$dimensional groups $\mathbf{f}_j^{(\tau)}=(\mathbf{f}_{j,1}^{(\tau)},...,\mathbf{f}_{j,k}^{(\tau)},...,\mathbf{f}_{j,c/d}^{(\tau)})$ where $\mathbf{f}_{j,k}^{(\tau)}\in\mathbb{R}^d$ specifies the $k^{th}$ group. For each group $\mathbf{f}_{j,k}^{(\tau)}$, the number of neighbors $N_i^{(t)}$ is sufficiently large to solve the following weighted least-squared approximation:
\begin{equation}
  A_{i,k}^*,b_{i,k}^*=\mathop{arg\,min}_{A,b}\sum_{{p}_j^{(\tau)}\in N_i^{(t)}}w_{j,k}^{(\tau)}||A\mathbf{f}_{j,k}^{(\tau)}+b-\tau||^2\,,
\end{equation}
where $w_{j,k}^{(\tau)}$ is the point-wise weight of each neighbor. The vanilla iterative normal refinement leverages weighted-PCA~\cite{66f8394a37054cab87290efb8ac34303} to solve the normals, which is unfriendly to back-propagation~\cite{wang2019backpropagation}. To this end, we attempt to directly fit this equation via its closed-form least-squared solution:
\begin{equation}
  [A_{i,k}^{T*},b_{i,k}^*]=({F}_{i,k}^{(t)T}{W}_{i,k}^{(t)}{F}_{i,k}^{(t)})^{-1}{F}_{i,k}^{(t)T}{W}_{i,k}^{(t)}\pmb{\tau}_{i,k}^{(t)}\,,
\end{equation}
where the weight matrix ${W}_{i,k}^{(t)}=diag(w_{1,k}^{(\tau)}, ..., w_{|N_i^{(t)}|,k}^{(\tau)})\in\mathbb{R}^{|N_i^{(t)}|\times|N_i^{(t)}|}$, the feature matrix ${F}_{i,k}^{(t)}=[(\mathbf{f}_{1,k},1),..,(\mathbf{f}_{|N_i^{(t)}|,k},1)]\in\mathbb{R}^{|N_i^{(t)}|\times(d+1)}$ and the time vector $\pmb{\tau}_{i,k}^{(t)}\in\mathbb{R}^{|N_i^{(t)}|}$. The normal vector is as follows:
\begin{equation}
  \mathbf{n}_{i,k}^{(t)}=\frac{(A_k^{T*},-1)}{||(A_k^{T*},-1)||}\,,
\end{equation}
where $||\cdot||$ is the $\ell 2$-norm. The concatenated normals $\mathbf{n}_{i}^{(t)}=concat(\mathbf{n}_{i,1}^{(t)},..,\mathbf{n}_{j,\frac{c}{d}}^{(t)})$ are fed into a 1$\times$1 convolution $C_l$ to obtain the $d_l-$dimensional abstracted dynamic feature $C_l(\mathbf{n}_{i}^{(t)})$. After the dimension alignment with another 1$\times$1 convolution $R_l$, the feature is forwarded to the next layer via a residual connection.
\begin{figure*}[h]
\captionsetup[subfloat]{justification=centering, skip=8pt}
  \centering\subfloat[{Ratio of $c$}]{
  \centering\includegraphics[width=0.22\textwidth]{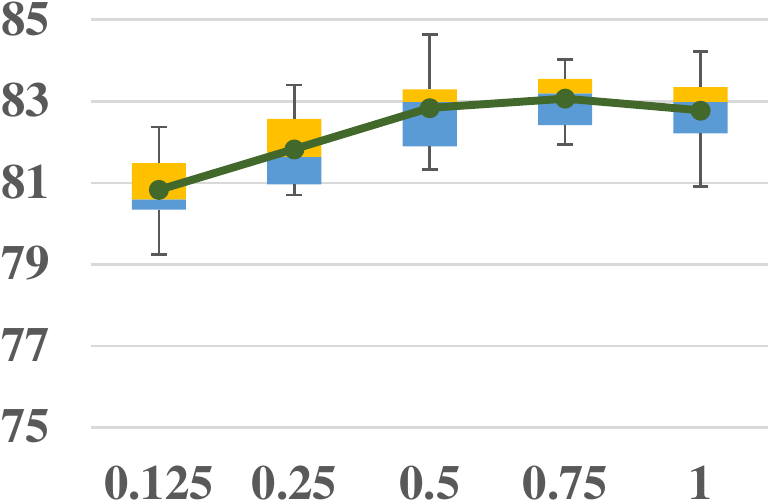}\label{fig:ratio_c}
  }~~~~
  \centering\subfloat[{Group Size $d$}]{
    \centering\includegraphics[width=0.22\textwidth]{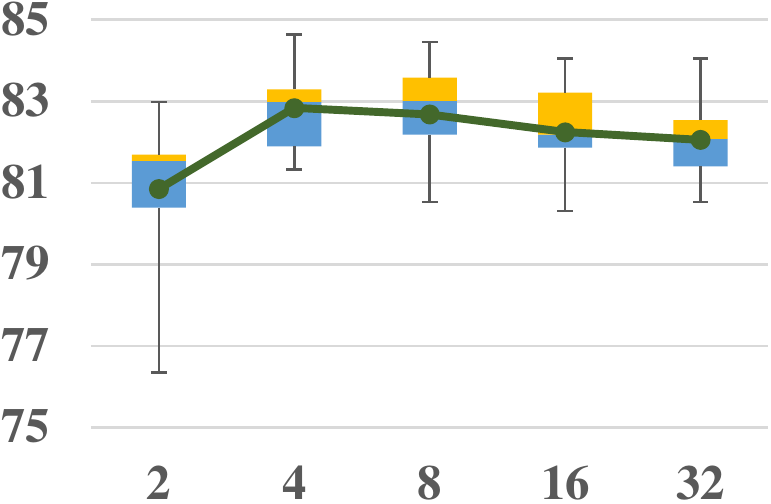}\label{fig:groupsize_d}
  }~~~~
  \centering\subfloat[{Temporal Radius $\Delta t$}]{
  \centering\includegraphics[width=0.22\textwidth]{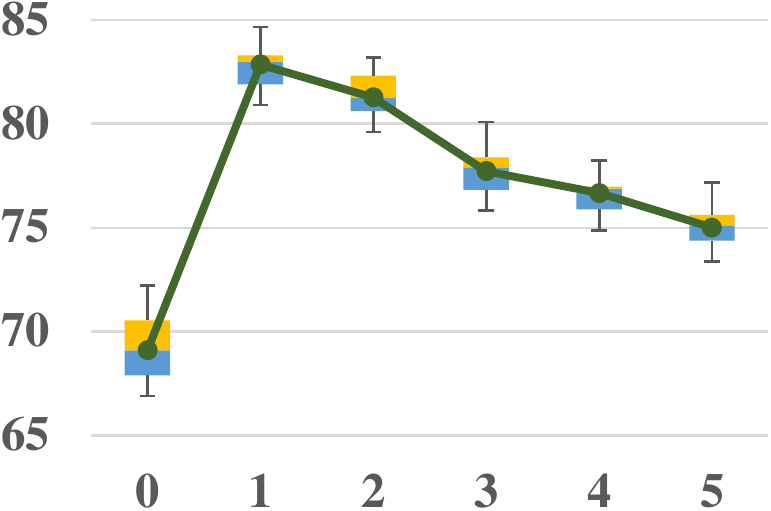}\label{fig:temp_dt}
  }~~~~
  \centering\subfloat[{Spatial Radius $\Delta r$}]{
  \centering\includegraphics[width=0.22\textwidth]{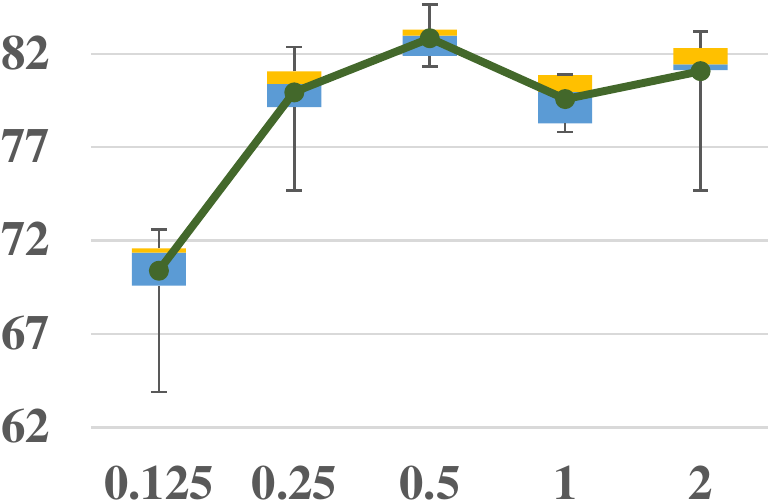}\label{fig:spat_dr}
  }
  \caption{\textit{Box-whisker plots of temporal-stream performance on validation set of \emph{NvGesture} under different hyper-settings.} The x-axis is the value of hyper-parameters, while y-axis is the validation accuracy (\%).}%
  \label{fig:hypersetting} 
\end{figure*}
\paragraph{Grouping with Weighted Neighbors (\textcolor{ForestGreen}{Green Dotted Frame} in Figure~\ref{fig:framework})} For each neighbor $p_j^{(\tau)}$, we compute its weight $w_{j,k}^{(\tau)}$ for better point-wise representations in the $l^{th}$ layer. Inspired by iterative normal refinement, we feed the channel-wise fitting deviation (a.k.a. residual) from the $(l-1)^{th}$ layer into a 1$\times$1 convolution $H_{l-1}^{k}$ activated by $sigmoid$ to obtain the neighbor weights within $[0,1]$:
\begin{equation}
  {w}_{j,k}^{(t)}=H_{l-1}^{k}(\mathbf{d}_{l-1}(p_j^{(\tau)}))\,,
\end{equation}
where $\mathbf{d}_{l-1}(p_j^{(\tau)})$ is the vector of fitting deviations in the $(l-1)^{th}$ layer. In the first layer, ${w}_{j,k}^{(t)}$ is set to $1$ for all of the neighbors. 
\subsubsection{Aggregation}
In the last layer, we leverage the dynamic features to obtain $softmax$ classification scores of a point cloud sequence. The spatial and temporal stream is respectively optimized through a cross-entropy loss in the training stage. During the testing phase, category predictions from the static model and the temporal stream are averaged as the final outputs. Since all the operations of kinematic representation learning are differentiable, it can be seamlessly plugged into a wide range of static neural architectures with minor structural surgery.

\section{Experiments}\label{sec:experiment}
To evaluate the performance of Kinet, we conduct experiments on three datasets (NvGesture~\cite{r3dcnn}, SHREC'17~\cite{de2017shrec}, MSRAction-3D~\cite{MSRAction-3D} and NTU-RGBD~\cite{shahroudy2016ntu}) for gesture recognition or action classification. The proposed framework is implemented with TensorFlow~\cite{tensorflow2015-whitepaper}. All experiments are conducted on the NVIDIA DGX-1 stations with Tesla V100 GPUs. In most experiments, PointNet++~\cite{qi2017pointnet++} is adopted as the static backbone in Kinet. For a fair comparison, we follow the identical settings of network layers to~\cite{min2019flickernet}. The other backbones are mainly chosen to evaluate the versatility. If not specified explicitly, we keep hyper-parameters of the original backbone (see \textbf{Appendix} for more details).  For training stability, we first train the static backbone (spatial stream) until convergence and then freeze its weights to individually optimize the dynamic branch (temporal stream). Following ~\cite{efficient_pointlstm,fan2021pstnet,liu2019meteornet}, classification accuracy is the main evaluation metric and we use default data splits for fair comparisons.
\subsection{NvGesture: Hyper-settings \& Ablations}
\textit{NvGesture} consists of 1532 (1050 for training and 482 for testing) videos composed of 25 classes. Following~\cite{efficient_pointlstm}, we uniformly sample 32 frames from a video and generate 512 points for each frame. To provide intuition behind the operation of our framework, we first investigate the impact of various network settings and components before we turn to further studies.

\noindent\textbf{Influence of Hyper-settings} We explore the effectiveness of crucial hyper-parameters with the 10-fold validation protocol~\cite{min2019flickernet}. To exclude the interference of the static branch, we calculate the accuracy only using the classification results of the dynamic stream without ensembles. Box-whisker plots are utilized to analyze the performance and the stability of hyper-settings as shown in Figure~\ref{fig:hypersetting}.

\textit{Feature dimensions} $c$ controls the output size of convolution $D_l$ to reduce computational overheads, as depicted in Figure~\ref{fig:framework}, which is proportional to the dimension of static features. We vary the ratio from $12.5\%$ to $100\%$ to find a value that makes the representations as compact as possible but informative enough. As shown in Figure~\ref{fig:ratio_c}, $50\%$ is adequate to encode the dynamic information from a static layer. In comparison with the standalone flow-based branch, this saves at least half of the network parameters.

\textit{Group size} $d$ controls the dimensions in each group. Ideally, $d$ is expected to be sufficiently large to describe motions. However, excessively large matrices cause inefficiency in the ST-normal solvers. By changing $d$ from $2$ to $32$, we empirically find that $d=4$ properly balances the accuracy and computational overheads as shown in Figure~\ref{fig:groupsize_d}.

\textit{Temporal radius} $\Delta t$ controls the receptive field of our dynamic branch along the temporal axis. It should be sufficiently small to retain details but large enough to model long-term interactions. It is observed that multi-frame information ($\Delta t\leq 1$) indeed improves the performance over a single frame ($\Delta t=0$). In the searching range of $[0,5]$, the best temporal radius $\Delta t$ is $1$ as depicted in Figure~\ref{fig:temp_dt}. 

\textit{Spatial radius} $\Delta r$ controls the receptive field in the spatial dimension. Similar to $\Delta t$, $\Delta r$ should have a moderate optimum. From the comparison of $\Delta r\in [0.125,2.0]$ in Figure~\ref{fig:spat_dr}, $\Delta r=0.5$ consistently has decent performance.

In the remaining experiments, we set the ratio of feature reduction as $50\%$, group-wise dimensions $d=4$, temporal radius $\Delta t=1$, and spatial radius $\Delta r=0.5$, respectively.

\begin{table}[h!]
\centering
\resizebox{0.48\textwidth}{!}{
\small\begin{tabular}{@{\extracolsep{4pt}}cccccc@{}}
\toprule[0.75pt]
\textbf{Settings}&\textbf{Spatial (Static)}   & \multicolumn{3}{c}{\textbf{Temporal (Dynamic)}} & \textbf{Accuracy}                \\ \cline{2-2} \cline{3-5}
\textbf{No.}&Pretrain & Pretrain & ST-normal       & Weighted      & \multicolumn{1}{c}{~~\textbf{(\%)}} \\ 
\toprule[0.75pt]
\textit{\romannumeral1.}&\xmark & \textbf{--}       & \textbf{--}       & \textbf{--}       & 82.6 \\ 
\textit{\romannumeral2.}&\cmark  & \textbf{--}       & \textbf{--}       & \textbf{--}       & 84.5 \\ 
\textit{\romannumeral3.}&\textbf{--}       & \xmark & \cmark  & \cmark  & 80.9 \\ 
\textit{\romannumeral4.}&\textbf{--}       & \cmark  & \cmark  & \cmark  & 82.4 \\\hline
\textit{\romannumeral5.}&\cmark  & \cmark  & \xmark & \xmark & 85.3 \\ 
\textit{\romannumeral6.}&\cmark  & \cmark  & \cmark  & \xmark & 87.9 \\ 
\textit{\romannumeral7.}&\cmark  & \cmark  & \cmark  & \cmark  & 89.1 \\
\toprule[0.75pt]
\end{tabular}}
\small\centering\caption{\textit{Ablation studies on \emph{NvGesture}.} {\cmark}/{\xmark} means that the operation is applied/not applied to the framework, while \textbf{--} means that the predictions of the corresponding spatial/temporal stream are excluded from the evaluation. The results of the upper/lower part is obtained from a single stream/two streams, respectively.}\label{tab:ablationresults}
\end{table}
\noindent\textbf{Ablation Studies} We conduct ablation studies on different components on the test set of \textit{NvGesture}.

\textit{Is it beneficial to pretrain on static datasets?} For two-stream models of image-based videos, it is well-known that pretraining on images significantly improves the performance on videos~\cite{NIPS2014_twostream,Carreira_2017_i3d}. However, this fact has not been verified for point cloud sequences. By pretraining the static PointNet++ on ModelNet40~\cite{chang2015shapenet}, we analyze the individual performance change for each branch. Table~\ref{tab:ablationresults} \textit{\romannumeral1.} \& \textit{\romannumeral2.} demonstrate that the performance of spatial stream on multi-frame predictions increases by $1.9\%$ (from $82.6\%$ to $84.5\%$), while the temporal branch using the pretrained static features also boosts the accuracy by $1.5\%$ as shown in Table~\ref{tab:ablationresults} \textit{\romannumeral3.} \& \textit{\romannumeral4.}. Obviously, both the static branch and the dynamic stream benefit from pretraining. Based on the pretrained backbone, the fusion of two-stream results achieves the accuracy of $89.1\%$ since the two branches are complementary to each other.

\textit{Is it useful to abstract with normal estimation?} To evaluate the efficacy of our abstracting operation, we remove the process of normal calculation in Figure~\ref{fig:framework}. Instead, the reduced feature is directly fed into the convolution $C_l$ and aggregated with max pooling. In this manner, all the trainable convolutions are unchanged but the accuracy significantly drops from $89.1\%$ to $85.3\%$ as shown in Table~\ref{tab:ablationresults} \textit{\romannumeral5.} \& \textit{\romannumeral7.}. By comparing \textit{\romannumeral2.} with \textit{\romannumeral5.} in Table~\ref{tab:ablationresults}, we find that purely introducing additional convolution parameters over the static backbone contributes only $0.8\%$ to the performance gain. Evidently, normal estimation is a vital component in the abstracting operation. 

\textit{Is it helpful to group with weighted neighbors?} By replacing all the learnable weights of neighbor points with the fixed value of $1.0$, we train the model to analyze the effect of weighted grouping. Table~\ref{tab:ablationresults} \textit{\romannumeral6.} shows that the performance decreases to $87.9\%$, which means the grouping operation with weighted neighbors is capable of further improving the performance of our feature abstraction.

\begin{table}[!h]
      \centering
    \scriptsize\begin{tabular}{rcc}
    \toprule[0.75pt]
    \textbf{Methods} & \textbf{Modalities}& \textbf{Accuracy (\%)} \\ %
    \toprule[0.75pt]
        R3DCNN~\cite{r3dcnn} & Infrared Image & 63.5 \\ 
        \hdashline
        R3DCNN~\cite{r3dcnn} & Optical Flow & 77.8 \\
        \hdashline  
        R3DCNN~\cite{r3dcnn} & Depth Map & 80.3 \\ 
        PreRNN~\cite{prernn} & Depth Map & 84.4 \\ 
        MTUT~\cite{mtut} & Depth Map & 84.9 \\
        \hdashline 
        R3DCNN~\cite{r3dcnn} & RGB Frame & 74.1 \\ 
        PreRNN~\cite{prernn} & RGB Frame & 76.5 \\ 
        MTUT~\cite{mtut} & RGB Frame & 81.3 \\
        \hdashline
        PointNet++~\cite{qi2017pointnet++} & Point Cloud & 63.9 \\ 
        FlickerNet~\cite{min2019flickernet} & Point Cloud & 86.3 \\ 
        PointLSTM-base~\cite{efficient_pointlstm} & Point Cloud & 85.9 \\ 
        PointLSTM-early~\cite{efficient_pointlstm} & Point Cloud & 87.9 \\ 
        PointLSTM-PSS~\cite{efficient_pointlstm} & Point Cloud & 87.3 \\ 
        PointLSTM-middle~\cite{efficient_pointlstm} & Point Cloud  & 86.9 \\
        PointLSTM-late~\cite{efficient_pointlstm} & Point Cloud & 87.5 \\
        Human~\cite{r3dcnn} & RGB Frame & 88.4 \\ 
        \hdashline
        \textbf{Kinet} & Point Cloud & \textbf{89.1} \\
    \toprule[0.75pt]
        \end{tabular}
    \small\caption{\textit{Quantatitve results achieved on \emph{NvGesture}. }}\label{tab:nvgest}
\end{table}
\noindent\textbf{Comparisons} \textit{NvGesture} is a multi-modality dataset and allows us to compare our method to state-of-the-art techniques under the standard data split~\cite{r3dcnn}. As shown in Table~\ref{tab:nvgest}, our approach with an accuracy of $89.1\%$ not only outperforms all of the existing point cloud methods, but also achieves higher performance than those models using other modalities. It is worth mentioning that Kinet is even superior to the human recognition on RGB videos ($88.4\%$) \textit{for the first time}.
\begin{figure*}
\captionsetup[subfloat]{skip=6pt}
  \centering
  \subfloat[{Depth, w/ BBox}]{
    \animategraphics[loop,autoplay,poster=7,width=0.14\textwidth]{8}{figure/nobk_depth/0}{0}{7}\label{fig:depth_bbox}
    }~~~~~~~~
  \subfloat[{Spatial, w/ BBox}]{
    \animategraphics[loop,autoplay,poster=7,width=0.14\textwidth]{8}{figure/shrec_static/0}{0}{7}\label{fig:spat_bbox}
    }~~~~~~~~
  \subfloat[{Temporal, w/ BBox}]{
  	\animategraphics[loop,autoplay,poster=7,width=0.14\textwidth]{8}{figure/shrec_flow/0}{0}{7}\label{fig:temp_bbox}
  }~~~~~~~~
  \subfloat[{Performance, w/ BBox}]{
    \includegraphics[width=0.2\textwidth]{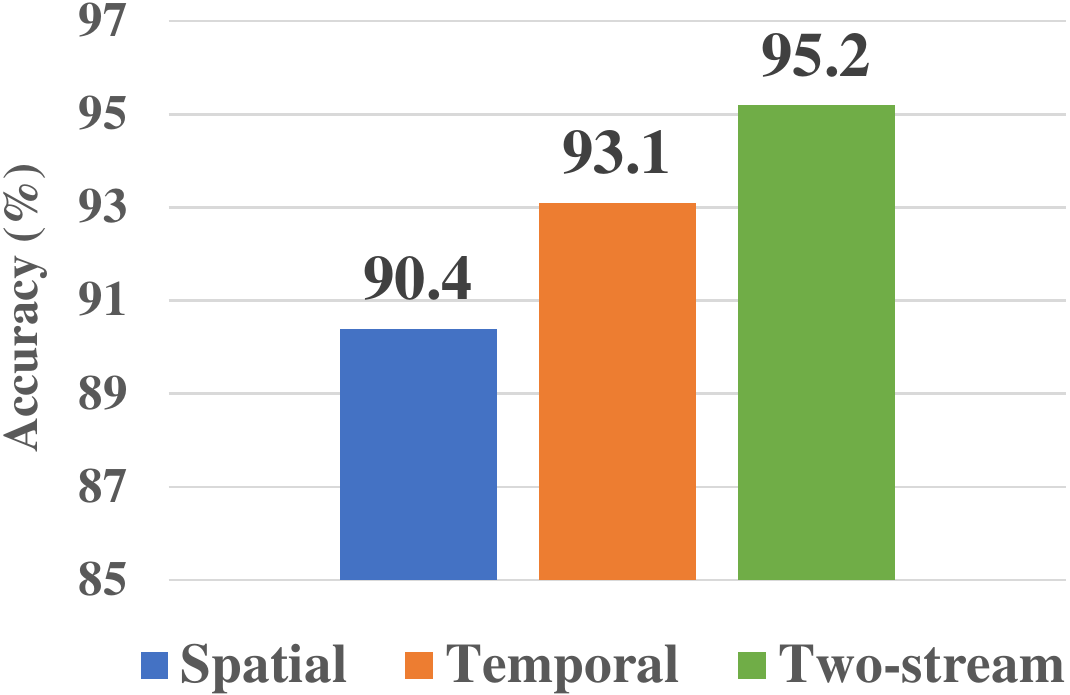}\label{fig:perf_bbox}
    }
    
  \subfloat[{Depth, w/o BBox}]{
    \animategraphics[loop,autoplay,poster=7,width=0.14\textwidth]{8}{figure/bk_depth/0}{0}{7}\label{fig:depth_nobbox}
    }~~~~~~~~
  \subfloat[{Spatial, w/o BBox}]{
    \animategraphics[loop,autoplay,poster=7,width=0.14\textwidth]{8}{figure/bk_shrec_static/0}{0}{7}\label{fig:spat_nobbox}
    }~~~~~~~~
  \subfloat[{Temporal, w/o BBox}]{
  	\animategraphics[loop,autoplay,poster=7,width=0.14\textwidth]{8}{figure/bk_shrec_flow/0}{0}{7}\label{fig:temp_nobbox}
  }~~~~~~~~
  \subfloat[{Performance, w/o BBox}]{
    \includegraphics[width=0.2\textwidth]{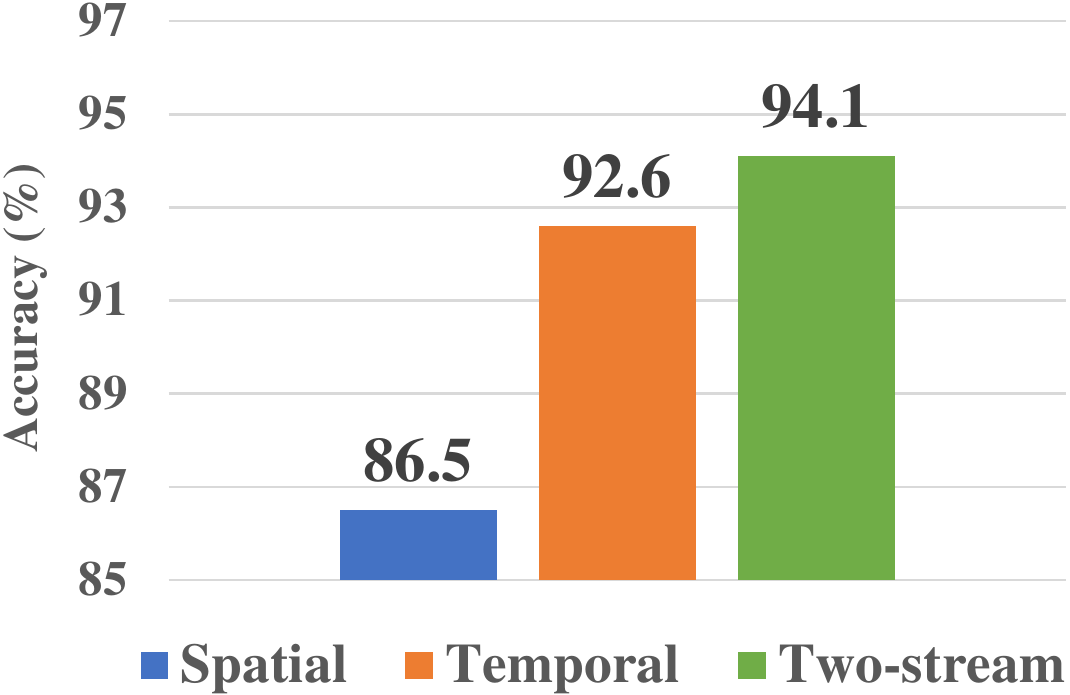}\label{fig:perf_nobbox}
    }
  \caption{\textit{Raw depth inputs, PACs and stream-wise accuracy on \emph{SHREC'17}.} In PACs, the points in \textcolor{red}{red} have the highest activation values, while the \textcolor{blue}{blue} ones are the lowest activating points. {\textcolor{cyan}{Best viewed in Adobe Reader where (a)-(c) \& (e)-(g) should play as videos.}}}\vspace{-4pt}
  \label{fig:visulization} 
\end{figure*}
\subsection{SHREC'17: Robustness to Noisy Backgrounds}
\textit{SHREC'17} is comprised of 2800 videos in 28 classes for gesture recognition, of which $70\%$ (2960 videos) are training data and the other $30\%$ (840 videos) are the test set. It has two types of supervisory signals, \ie, video-level classification labels and bounding boxes (BBox) of hand skeletons. Unlike most of the prior works only focusing on the background-free cases, we adopt two input settings to verify whether Kinet can capture useful movements and ignore meaningless ones:  1) w/ BBox (Figure~\ref{fig:depth_bbox}) - used by a majority of existing methods for high accuracy, based on the area inside the bounding boxes of hand skeletons without background interference; 2) w/o BBox (Figure~\ref{fig:depth_nobbox}) - raw videos with noisy backgrounds (the performer's body).

\noindent\textbf{Qualitative Analysis on Robustness} We visualize the learned point activation clouds (PACs)~\cite{min2019flickernet,fan2021pstnet} in Figure~\ref{fig:visulization}. With bounding boxes removing noisy backgrounds, the two streams work complementary - the static branch (Figure~\ref{fig:spat_bbox}) highlights the main parts (the palms) of spatial appearances, whereas the temporal representations (Figure~\ref{fig:temp_bbox}) capture key motions, such as {the movement of fingers and wrists}. In the case with redundant backgrounds (without bounding boxes), the static stream (Figure~\ref{fig:spat_nobbox}) excessively focuses on the large yet useless background portions (the performer's body), while the temporal stream (Figure~\ref{fig:temp_nobbox}) captures the moving parts (arms and fingers). Inevitably, the temporal stream also highlights several redundant points of the performer's shaking head by mistake. 

\noindent\textbf{Quantitative Analysis on Robustness} From Figure~\ref{fig:perf_bbox} \& \ref{fig:perf_nobbox}, it is observed that the dynamic branch shows strong robustness to motion-irrelevant backgrounds, where the accuracy slightly drops from $93.1\%$ to $92.6\%$, compared with that of the spatial stream which plunges by nearly $4\%$.

\noindent\textbf{Comparison} As shown in Table~\ref{tab:SHREC}, we compare the performance of Kinet to existing models. With the same bounding box supervision signal, our two-stream PointNet++ outperforms others with the accuracy of $95.2\%$. Noticeably, even for the highly challenging inputs without bounding boxes, our framework boosts the accuracy of static PointNet++ from only $86.5\%$ to $94.1\%$, which is comparable to state-of-the-art models with bounding boxes.
\begin{table}[t]
\begin{center}
      \centering
       \scriptsize\begin{tabular}{rccc}
    \toprule[0.75pt]
    \textbf{Methods} & \textbf{Modalities}& \textbf{BBox}& \textbf{Accuracy (\%)} \\ %
    \toprule[0.75pt]
            Key frames~\cite{de2017shrec} & Depth Map &\xmark & 71.9 \\
            \hdashline  
        SoCJ+HoHD+HoWR~\cite{socj} & Skeleton & \cmark& 81.9 \\ 
        Res-TCN~\cite{res-tcn} & Skeleton & \cmark& 87.3 \\ 
        STA-Res-TCN~\cite{res-tcn} & Skeleton & \cmark& 90.7 \\ 
        ST-GCN~\cite{ST_GCN} & Skeleton & \cmark& 87.7 \\ 
        DG-STA~\cite{DGSTA} & Skeleton & \cmark& 90.7 \\\hdashline
        PointLSTM-base~\cite{efficient_pointlstm} & Point Cloud &\cmark  & 87.6 \\ 
        PointLSTM-early~\cite{efficient_pointlstm} & Point Cloud &\cmark & 93.5 \\ 
        PointLSTM-PSS~\cite{efficient_pointlstm} & Point Cloud &\cmark   & 93.1 \\ 
        PointLSTM-middle~\cite{efficient_pointlstm} & Point Cloud &\cmark& 94.7 \\ 
        PointLSTM-late~\cite{efficient_pointlstm} & Point Cloud &\cmark  & 93.5 \\ 
        \hdashline
        \textbf{Kinet} & Point Cloud &\cmark&\textbf{95.2} \\
        \textbf{Kinet} & Point Cloud &\xmark& 94.1 \\\toprule[0.75pt]
      \end{tabular}
\end{center}
\small\centering\caption{\textit{Quantitative results achieved on \emph{SHREC'17}.}}\vspace{-6pt}\label{tab:SHREC}
\end{table}
\begin{table*}[!h]
\small
\begin{center}
\small\begin{tabular}{rccc}
\toprule[0.75pt]
\textbf{Methods} & \textbf{Modalities} & \textbf{\# of Frames} & \textbf{Accuracy (\%)} \\
\toprule[0.75pt]
Vieira \etal~\cite{vieira2012stop} & Depth Map & 20 & 78.20 \\
Kl\"{a}ser \etal~\cite{klaser2008spatio} & Depth Map & 18 & 81.43 \\
\hdashline
Actionlet~\cite{wang2012mining} & GroundTruth Skeleton & Full & 88.21 \\ 
\hdashline
PointNet++~\cite{qi2017pointnet++} & Point Cloud & 1 & 61.61  \\  \hdashline
MeteorNet~\cite{liu2019meteornet} & Point Cloud & 4 / 8 / 12 / 16 / 24 & 78.11 / 81.14 / 86.53 / 88.21 / 88.50\\
P4Transformer~\cite{fan21p4transformer} & Point Cloud & 4 / 8 / 12 / 16 / 24 & 80.13 / 83.17 / 87.54 / 89.56 / 90.94\\
PSTNet~\cite{fan2021pstnet} & Point Cloud & 4 / 8 / 12 / 16 / 24 & \textbf{81.14} / 83.50 / 87.88 / 89.90 / 91.20\\
\textbf{Kinet} & Point Cloud & 4 / 8 / 12 / 16 / 24 & 79.80 / \textbf{83.84} / \textbf{88.53} / \textbf{91.92} / \textbf{93.27}\\
\toprule[0.75pt]
\end{tabular}\vspace{-8pt}
\end{center}
\centering\caption{\vspace{-0.2cm}\textit{Quantitative results achieved on \emph{MSRAction-3D}.}}\label{tab:msra}\vspace{-5pt}
\end{table*}

\subsection{MSRAction-3D: Different Tasks \& Backbones}\label{sec:exp_msr}
\textit{MSRAction-3D} has 567 videos of 20 action categories. Following Fan~\etal~\cite{fan21p4transformer}, we adopt the standard data splitting protocol~\cite{liu2019meteornet, wang2012mining} and report the average accuracy over 10 runs.

\noindent\textbf{Versatility across Static Backbones} Under the taxonomy defined in~\cite{guo2020deep}, three types of backbones dominate the classification models of static point clouds: MLP-based, convolution-based, and graph-based methods. In this paper, we choose one typical static architecture for each paradigm respectively, \ie, MLP-based PointNet++~\cite{qi2017pointnet++}, convolution-based SpiderCNN~\cite{xu2018spidercnn} and graph-based DGCNN~\cite{dgcnn} in order to demonstrate the ease of extending these to dynamic point cloud tasks. By grouping a video into 16-frame clips as the input unit, we run 3 sets of experiments for each static model: 1) Directly feed videos into the static model; 2) Fuse the static spatial model and the temporal streams; 3) Ensemble classification scores from two static models, one is trained on the raw point clouds, while the other flow-based model is trained on scene flow estimated by Justgo~\cite{justgo}, a self-supervised scene flow estimation tool. Apart from classification performance, we also take memory consumption and computational complexity into consideration. They are measured with the accuracy, the number of parameters, and floating-point operations per second (FLOPS), respectively.

\begin{figure}[!t]
\centering
\includegraphics[width=0.95\linewidth]{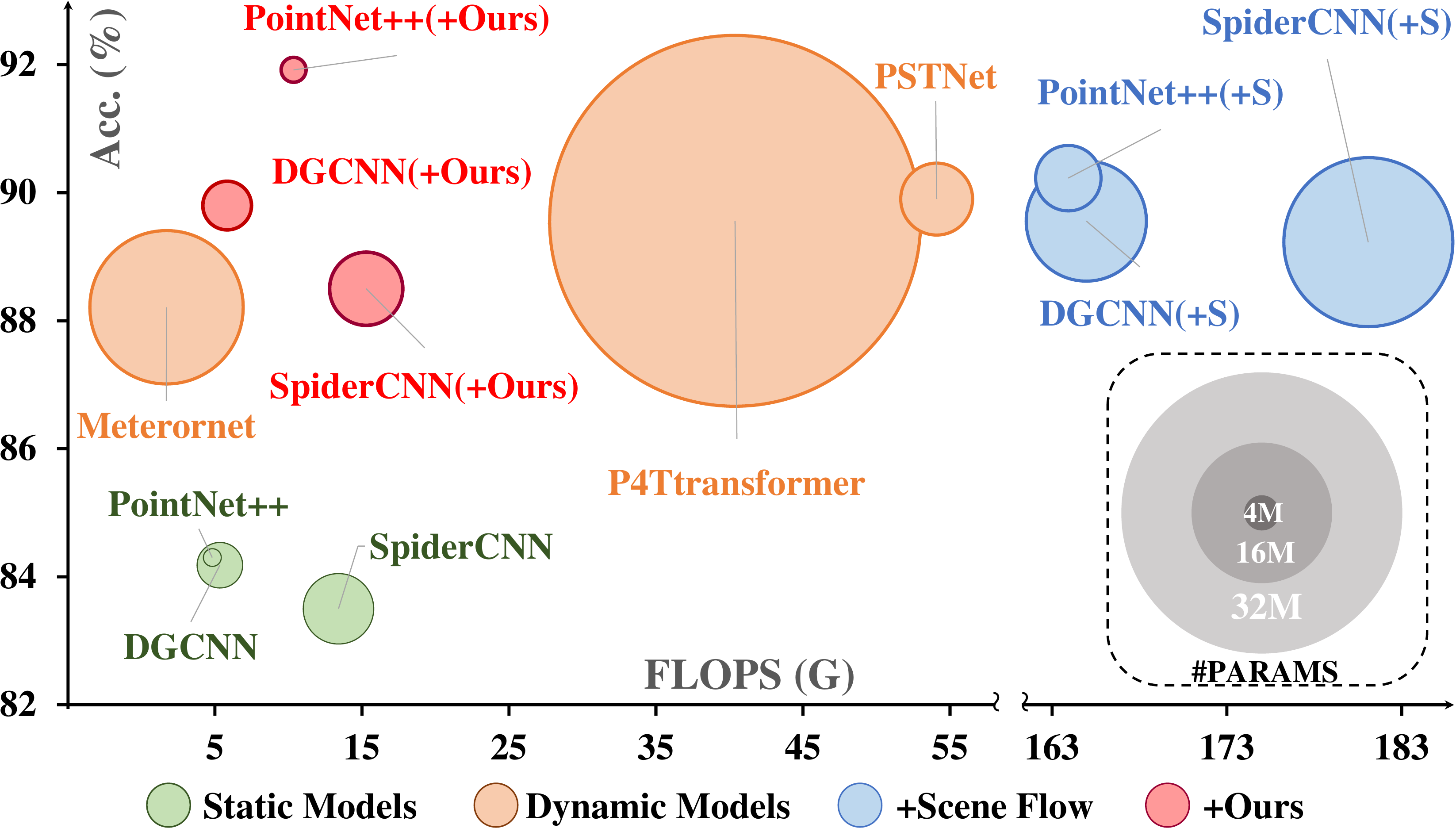}
    \caption{{\textit{Comparison of FLOPS, parameter number, and accuracy on 16-frame \emph{MSRAction-3D}.} Quantitative details can be found in \textbf{Appendix}.}\vspace{-0.6cm}}
    \label{fig:ball_chart}
\end{figure}
As illustrated in Figure~\ref{fig:ball_chart}, the extra input modality of estimated scene flow in setting 3) (\textit{+Scene Flow} \textcolor{Cerulean}{\FilledCircle}) considerably improves the accuracy of the three static models (\textcolor{YellowGreen}{\FilledCircle}) to a level comparable to the state-of-the-art. However, the scene flow estimator and another flow-based classifier almost triple the number of parameters. Even worse, the estimation of scene flow introduces more than 150G FLOPS of extra calculations since the point-wise dense predictions are required between every two consecutive frames. For setting 2) (\textit{+Ours} \textcolor{WildStrawberry}{\FilledCircle}), it is observed that our kinematic representations consistently increase the accuracy of the static predictions by 5.99\%$\sim$9.04\% relative gains. By utilizing the kinematic representations, the FLOPS only increases to 5.83G$\sim$15.29G, and the number of parameters increases by 0.59M$\sim$1.08M. These computing overheads are negligible and make the fused model extremely lightweight. Compared with state-of-the-art dynamic networks (\textcolor{Melon}{\FilledCircle}), they achieve comparable or superior performance with fewer model parameters and lower computational complexity. Intriguingly, Kinet has higher performance gains in higher-performance static models, possibly indicating its \textit{limitation} - the informativeness of kinematic representations is constrained by the static features. A poor static backbone 
would benefit from the proposed approach, but it cannot improve the underlying static representation.

\noindent\textbf{Comparison} We compare the PointNet++-based Kinet with existing classification models on dynamic point clouds. Following prior research~\cite{liu2019meteornet,fan2021pstnet,fan21p4transformer}, we set the length as $\{4,8,12,16,24\}$ frames (2048 points/frame) and report the mean accuracy of 10 runs. As shown in Table~\ref{tab:msra}, our framework shows the superiority in long videos (length$\geq 8$) and achieves the 24-frame accuracy of $93.27\%$, exceeding others by a large margin.
\subsection{NTU-RGBD: Effectiveness on Large-scale Data}
After we add the author list and the acknowledgement section to our camera-ready version, this part has to be moved to \textbf{Appendix} because of limited space. Please see \textbf{Appendix} for more details.
\section{Conclusion}\label{Conclusion}
In this paper, we propose Kinet to bridge the gap between static point cloud models and dynamic sequences. By extending kinematic ST-surfaces to the high-dimensional feature space and unrolling the ST-normal solver differentiably, the presented framework \underline{gains} advantages from mature static models. Without the \underline{pain} of modeling point-wise correspondences, it can be seamlessly integrated into arbitrary static point cloud learning backbones, with only minor structural surgery and low computing overhead. Experiments on four datasets, two tasks, and three static networks demonstrate the efficacy of our framework in dynamic classification, the efficiency in parameters and FLOPS, and the versatility to various static backbones. An obvious limitation is that Kinet is constrained by the performance of the static model - a poor static baseline will not be rectified by adding our dynamic extension. Therefore, we will explore a method to self-improve static representations in future works.
\paragraph{Acknowledgements} Our research is supported by Amazon Web Services in the Oxford-Singapore Human-Machine Collaboration Programme and by the ACE-OPS project (EP/S030832/1). We are grateful to the three anonymous reviewers for their valuable comments. 
{\small
\bibliographystyle{ieee_fullname}
\bibliography{egbib}
}


\section*{Supplementary material}

# No Pain, Big Gain: Classify Dynamic Point Cloud Sequences with Static Models by Fitting Feature-level Space-time Surfaces (Supplementary Materials)

Jia-Xing Zhong, Kaichen Zhou, Qingyong Hu[✉], Bing Wang, Niki Trigoni, Andrew Markham

Deparment of Computer Science, University of Oxford

{jiaxing.zhong, rui.zhou, qingyong.hu, bing.wang, niki.trigoni, andrew.markham}@cs.ox.ac.uk

## 1. NTU-RGBD: Effectiveness on Large-scale Data

*NTU-RGBD* has two subsets, *i.e.*, NTU-RGBD 60 [21] and NTU-RGBD 120 [9]. We evaluate our model on the former since it is included in more comparison results of previous researches. NTU-RGBD 60 is a large-scale dataset for 3D action recognition, with 56880 RGBD videos of 60 action categories. The data is converted into point cloud sequences with the implementation of PSTNet [4].

**Comparison** Following the official data partition [21], cross-subject and cross-view scenarios are individually adopted as two splits. As shown in Table 1, Kinet is superior to the others, with an accuracy of 92.3% in the cross-subject split. In the cross-view protocol, Kinet comes a close second (with 96.4%) after 96.5% of PSTNet.

| Methods | Modalities | Accuracy (%) | |
|---|---|---|---|
| | | Cross-subject | Cross-view |
| SkeleMotion [2] | Skeleton | 69.6 | 80.1 |
| GCA-LSTM [11] | Skeleton | 74.4 | 82.8 |
| Attention-LSTM [10] | Skeleton | 77.1 | 85.1 |
| AGC-LSTM [24] | Skeleton | 89.2 | 95.0 |
| AS-GCN [8] | Skeleton | 86.8 | 94.2 |
| VA-fusion [32] | Skeleton | 89.4 | 95.0 |
| 2s-AGCN [23] | Skeleton | 88.5 | 95.1 |
| DGNN [22] | Skeleton | 89.9 | 96.1 |
| MS-G3D [13] | Skeleton | 91.5 | 96.2 |
| HON4D [19] | Depth Map | 30.6 | 7.3 |
| SNV [31] | Depth Map | 31.8 | 13.6 |
| $HOG^2$ [18] | Depth Map | 32.2 | 22.3 |
| Li *et al*. [7] | Depth Map | 68.1 | 83.4 |
| Wang *et al*. [26] | Depth Map | 87.1 | 84.2 |
| MVDI [29] | Depth Map | 84.6 | 87.3 |
| 3DV-Appearance [28] | Point Cloud | 80.1 | 85.1 |
| 3DV-Motion [28] | Voxel | 84.5 | 95.4 |
| 3DV-Full [28] | Point Cloud + Voxel | 88.8 | 96.3 |
| P4Transfomers [3] | Point Cloud | 90.2 | 96.4 |
| PSTNet [4] | Point Cloud | 90.5 | **96.5** |
| **Kinet** | Point Cloud | **92.3** | 96.4 |

Table 1. *Quantitative results achieved on* NTU-RGBD 60.

## 2. Implementation Details

For the sake of reproducibility, we elaborate the implementation as detailed as possible and the source code will be released soon.

### 2.1. Matrix Inversion

Equation (6) in the main body of this paper is to fit the group-wise ST-surface via the closed-form least-squared solution:

$$[A_k^{T*}, b_k^*] = (F_{i,k}^{(t)T} W_{i,k}^{(t)} F_{i,k}^{(t)})^{-1} F_{i,k}^{(t)T} W_{i,k}^{(t)} \boldsymbol{\tau}_{i,k}^{(t)}, \quad (1)$$

where the weight matrix $W_{i,k}^{(t)} = diag(w_{1,k}^{(\tau)}, ..., w_{|N_i^{(t)}|,k}^{(\tau)}) \in \mathbb{R}^{|N_i^{(t)}| \times |N_i^{(t)}|}$, the feature matrix $F_{i,k}^{(t)} = [(\mathbf{f}_{1,k}, 1), ..., (\mathbf{f}_{|N_i^{(t)}|,k}, 1)] \in \mathbb{R}^{|N_i^{(t)}| \times (d+1)}$ and the time vector $\boldsymbol{\tau}_{i,k}^{(t)} \in \mathbb{R}^{|N_i^{(t)}|}$.

Note that we do not utilize the default implementation of matrix inversion on Tensorflow [1] since the default implementation using the LU decomposition is specially designed for large matrices. In our settings, matrices are relatively small in each group and we adopt home-brew approaches to matrix inversion. Given an $n \times n$ square matrix $M \in \mathbb{R}^{n \times n}$, we inverse the matrix according to its size.

In the case of extremely small matrices ($n \leq 4$), we directly inverse the matrix with element-wise calculations. For example, let $M = \begin{pmatrix} m_1 & m_2 \\ m_3 & m_4 \end{pmatrix} \in \mathbb{R}^{2 \times 2}$. The inversion of this matrix is $M^{-1} = \frac{1}{m_1 m_4 - m_2 m_3} \begin{pmatrix} m_4 & -m_2 \\ -m_3 & m_1 \end{pmatrix}$, where $m_1 m_4 - m_2 m_3$ is clipped to the minimal value of $1 \times 10^{-6}$ for the numerical stability of singular matrices. In terms of medium matrices ($4 < n \leq 16$), we recursively inverse the partitioned matrices based on the aforementioned element-wise operations for small matrices. For the case $n > 16$, we leverage Cholesky decomposition [6] to solve the matrix inversion.

### 2.2. Network Structures

#### 2.2.1 Kinet: MLP-based Backbone (PointNet++)

The MLP-based methods separately model every point with shared multi-layer perceptions (MLPs), followed by a symmetric aggregation (*e.g.*, max pooling) to fuse order-invariant information. In this paper, we choose PointNet++ [20] as our MLP-based backbone for static point clouds. For a fair comparison, we keep the number of parameters in each layer identical to FlickerNet [15], one of the state-of-the-art gesture recognition networks on sequential point clouds.

#### 2.2.2 Kinet: Graph-based Backbone (DGCNN)

Unlike the MLP-based backbones independently capturing point-level representations, graph-based methods model pointwise interactions by regarding a point cloud as a graph, of which each point is the vertex and edges is established upon the neighboring distribution of these points. For the graph-based paradigm, we conduct experiments on the static backbone of DGCNN [27] and keep the default settings of layer-wise parameters including the feature aggregation of the channel-wise additions in [27].

#### 2.2.3 Kinet: Conv.-based Backbone (SpiderCNN)

Different from the other two paradigms using graphs or MLPs, the conv.-based models directly devise the convolution particularly for unstructured point clouds. We adopt SpiderCNN [30] as our convolution-based static backbone and keep the same layer-wise settings including the feature aggregation of **concatenation** and **top-k pooling**.

### 2.3. Training Configurations

The proposed framework is implemented with Tensorflow [1]. All experiments are conducted on the NVIDIA DGX-1 stations with Tesla V100 GPUs. During the training stage, the hyper-parameters are $batch\_size = 16$, $base\_learning\_rate = 0.001$, with an Adam optimizer [5]. The number of training epochs depends on various datasets: 200 epochs on *NvGesture*/*SHREC'17*, 150 epochs on *MSRAction-3D*, and 20 epochs on *NTU-RGBD*. For training stability, we first train the static backbone (spatial stream) until convergence and then freeze its weights to individually optimize the dynamic branch (temporal stream). To make full use of the features in various scales, multiple layers are connected to output the final results. As for the hyper-settings of Kinet itself, we set the ratio of feature reduction as $50\%$, group-wise dimensions $d = 4$, temporal radius $\Delta t = 1$, and spatial radius $\Delta r = 0.5$, respectively.

## 2.4. Point Activation Clouds

This concept stems from FlickerNet [15], which indicates the highlighted points of a model. Likewise, P4Transformer [3] and PSTNet [4] have the similar visualization concepts of the attentional values and the convolutional outputs, respectively. For readability, we formulate the concept of Point Activation Clouds (PACs) for the proposed two-stream framework.

Given a frame $P_t$ of point clouds at the $t^{th}$ time step, denote the activated feature vector of the $i^{th}$ point $p_i^{(t)}$ in the $l^{th}$ layer as $\phi_l(p_i^{(t)})$. The PAC of a centroid $p_i^{(t)}$ in this layer is defined as:

$$PAC_l(p_i^{(t)}) = \max_{\phi_l \in \boldsymbol{\phi}_l} \max_{p_j^{(\tau)} \in N_i^{(t)}} \phi_l(p_j^{(\tau)}), \qquad (2)$$

where $N_i^{(t)}$ is the space-time neighbor set of $p_i^{(t)}$. Thus, the larger PACs reflect the greater activated values for these centroids, which highlights the discernible outputs in the $l^{th}$ layer.

In the main body of this paper, we visualize the PACs of the last layer before the first max-pooling operation in the static stream of PointNet++ [20] and the corresponding layer in our dynamic branch.

## 2.5. Point-wise Scene Flow Estimation of Flow-based Baselines

In Section 4.3 of the main body, a self-supervised framework *Justgo* [17] is trained to estimate scene flow and we adopt it as an additional input in the setting 3). Note that Kinet does not require scene flow and setting 3) serves as the flow-based baseline.

### 2.5.1 Self-supervised Training Details

Based on the model pre-trained on *Flythings3D* [14], we train the scene flow estimator with the following hyper-parameters: $batch\_size = 8$, $radius = 5$, $flip\_prob = 0.5$, $base\_learning\_rate = 0.001$, with an Adam optimizer [5]. The balancing weight of the nearest neighbor loss and the cycle loss is set as 1:1.

### 2.5.2 Inference

The scene flow is first extracted by applying the trained *Justgo* model. Then, we scale the estimated scene flow to the same range of the raw point clouds, so that a static model with default hyper-parametric settings can be directly applied to the input modality of scene flow. Note that the number of parameters in a trained *Justgo* (excluding the parameters of the Adam optimizer) achieves $3.54$M. For a 16-frame snippet with 2048 points per frame, the snippet-wise computational complexity of scene flow extraction is $154.29$G FLOPs. The computational overhead of scene flow estimation is considerable.

### 2.5.3 Visualization

As depicted in Figure 1b, scene flow is decently estimated between two consecutive frames (as depicted in Figure 1a) with the self-supervised network. It is observed that key motions of the runner's legs are well captured but the interpolation results for the runner's head are insufficiently accurate: scene flow estimation without explicit supervision is a highly challenging task. As one of the input modalities for a vanilla two-stream model, the scaled scene flow in Figure 1c highlights the legs' movements, still preserving crucial dynamic information.

## 3. Detailed Experimental Results

### 3.1. Quantitative Computational Costs

In section 4.3 of the main body, we conduct 3 groups of experiments: 1) Directly feed videos into the static model; 2) Fuse the static model and the dynamic branch; 3) Ensemble classification scores from two static models, one is trained on the raw point clouds, while the other is trained on estimated scene flow. As shown in Table 2, the extra input modality of estimated scene flow in setting 3) (+S, Flow-based Two Streams) considerably improves the accuracy of the three static models (Static PointNet++, DGCNN and SpiderCNN) to a level comparable to the state-of-the-art. However, the scene flow estimator and another flow-based classifier almost triple the number of parameters. Even worse, the estimation of scene flow introduces more than 150G FLOPS of extra calculations since the point-wise dense predictions are required between every two consecutive frames. For setting 2) (+Ours, Kinet), it is observed that our kinematic representations consistently increase

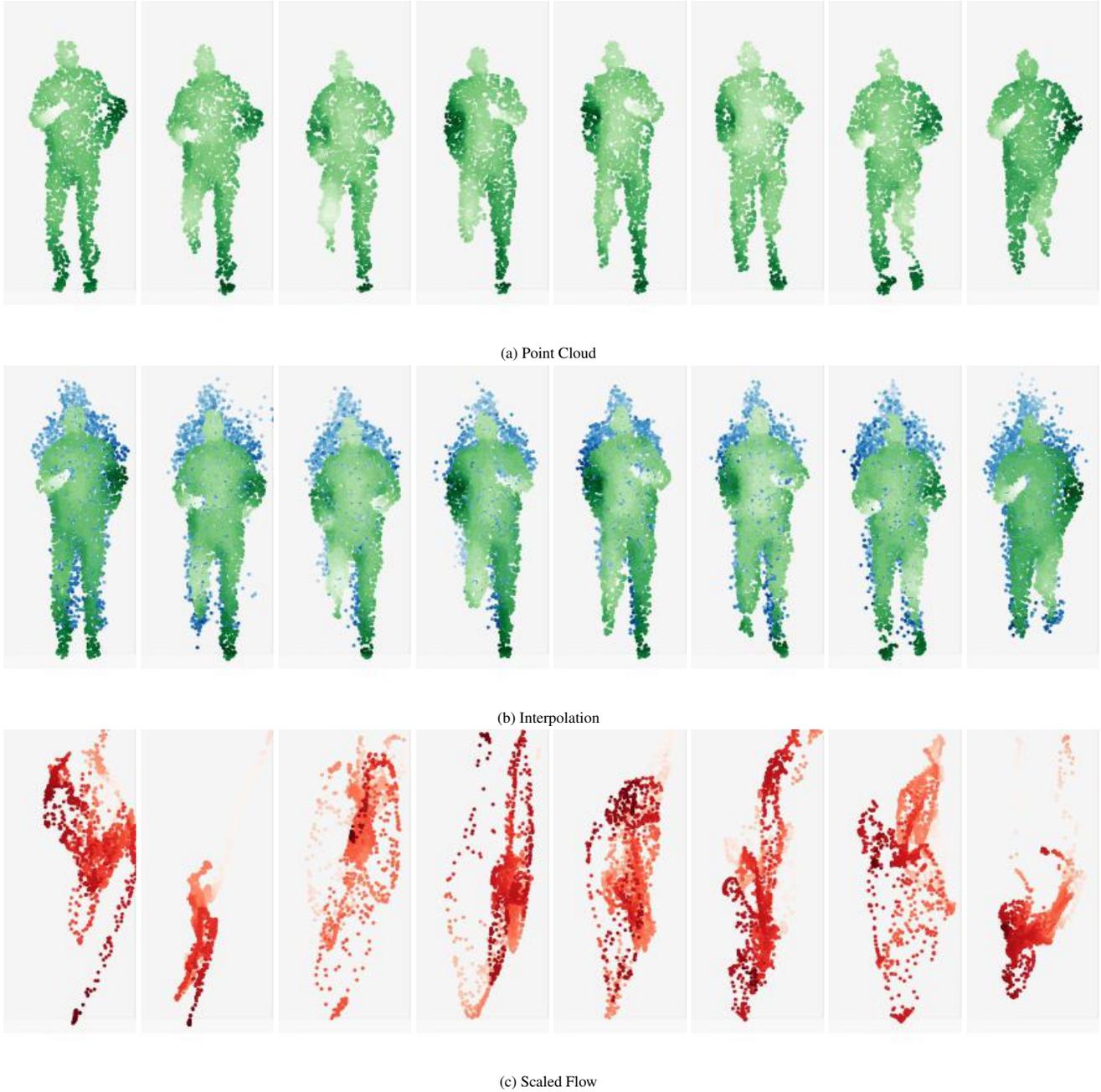

Figure 1. *Estimated point-wise scene flow of the flow-based baselines on* MSRAction-3D. The darker color indicates the greater depths (in point clouds) or the larger motions (in scene flow). (a) is the input of raw point clouds. For the frame interpolation in (b), the green points are the ground-truth point clouds, while the blue points are the interpolation results based on the last frame and the estimated scene flow. For the estimated scene flow in (c), it is normalized to the same scale as raw point clouds.

the accuracy of the static predictions by 5.99%∼9.04% relative gains. By utilizing the kinematic representations, the FLOPS only increases to 5.83G∼15.29G, and the number of parameters increases by 0.59M∼1.08M. These computing overheads are negligible and make the fused model extremely lightweight.

| Model | FLOPS (G) | #PARAMS (M) | Accuracy (%) |
|---:|---:|---:|---:|
| Static PointNet++ [20] | 4.82 | 2.13 | 84.30 |
| Flow-based Two Streams (PointNet++) | 163.93 | 7.79 | 90.23 |
| **Kinet (Pointnet++)** | 10.35 | 3.20 | 91.92 |
| Static DGCNN [27] | 5.33 | 5.25 | 84.18 |
| Flow-based Two Streams (DGCNN) | 164.95 | 14.04 | 89.56 |
| **Kinet (DGCNN)** | 5.83 | 5.85 | 89.82 |
| Static SpiderCNN [30] | 13.40 | 8.03 | 83.49 |
| Flow-based Two Streams (SpiderCNN) | 181.09 | 19.60 | 89.23 |
| **Kinet (SpiderCNN)** | 15.29 | 8.62 | 88.54 |
| Meterornet [12] | 1.70 | 17.60 | 88.21 |
| PSTNet [4] | 54.09 | 8.44 | 89.90 |
| P4Ttransformer [3] | 40.38 | 42.07 | 89.56 |

Table 2. *Quatative results of parameter number, FLOPS and accuracy on 16-frame* MSRAction-3D.

## 3.2. Per-class Performance Gains

Following Liu *et al.* [12], we report the performance gains over the original static model on *MSRAction-3D*. Under the same experimental configurations as the main body, we take 16 frames as an input unit and sample 2048 points for each frame. By comparing with the baseline (*PointNet++*) accuracy of setting 1), we report the performance changes of setting 2) and 3) in per-class action categorization.

As illustrated in Figure 4a, our dynamic branch significantly improves the classification accuracy of the static model. Noticeably, the categories of "High Throw" and "Hand Catch" show more than 30% absolute performance gains and most of the other performance changes achieve at least 20% improvements. There is only one category ("Horizontal Arm Wave") with about 6% performance decline. Figure 4b demonstrates that the extra input modality of estimated scene flow also boosts the overall performance. It has a large performance drop of more than 10% in the category of "High Arm Wave" and limited performance gains on many classes. Compared with the raw scene flow, our kinematic approach has high robustness and considerable improvements because it does not rely on the inaccurate estimation of scene flow.

## 3.3. Confusion Matrices

Following prior researches [15, 16] on point-based gesture recognition, we utilize confusion matrices to report the detailed performance on *NvGesture* and *SHREC'17*. A confusion matrix (a.k.a. an error matrix) shows whether a classifier is confounded by two categories, of which each column represents the instances in an actual class and each row represents the examples in a predicted category.

In the main body, we evaluate our model on *SHREC'17* under two input cases: 1) based on the entire video with backgrounds of the performer's body (w/o BBox); 2) based on the area of hand skeletons inside the bounding boxes (w/ BBox). It is observed in Figure 5 that the absence of bounding boxes leads to misclassification for more categories (*e.g.*, "Tap-1" and "Swipe Down-1") but the total number of error instances is quite small. From this, it is observed that Kinet manifests high robustness to backgrounds.

As depicted in Figure 6, our framework is able to distinguish an overwhelming majority of gestures. However, it fails in some extremely challenging cases. For example, the classifier confuses the category "Show Two Fingers" with "Push Two Fingers Away" possibly because point clouds are scarce and sparse in the part of fingers, which makes it difficult to capture subtle differences between these two gestures.

## 4. More Visualizations

### 4.1. Point-level Sequential PACs

Due to the page limitation in the main body, we visualize the depth videos and dynamic PACs in Figure 6 with the format of animations. In case the PDF reader cannot display the animations normally, we provide the sequential images as shown in Figure 2.

Unlike most of the prior works only focusing on the background-free cases, we adopt two input settings to verify whether Kinet can capture useful movements and ignore meaningless ones: 1) w/ BBox (Figure 2a) - used by a majority of existing

methods for high accuracy, based on the area inside the bounding boxes of hand skeletons without background interference; 2) w/o BBox (Figure 2d) - raw videos with noisy backgrounds (the performer's body). With bounding boxes removing noisy backgrounds, the two streams work complementary - the static branch (Figure 2b) highlights the main parts (the palms) of spatial appearances, whereas the temporal representations (Figure 2c) capture key motions, such as the movement of fingers and wrists. In the case with redundant backgrounds (without bounding boxes), the static stream (Figure 2e) excessively focuses on the large yet useless background portions (the performer's body), while the temporal stream (Figure 2f) captures the moving parts (arms and fingers). Inevitably, the temporal stream also highlights several redundant points of the performer's shaking head by mistake.

### 4.2. Video-level t-SNE Features

Different from PACs encoding point-level activation in intermediate layers, the last-layer features before the classifier aggregate the video-level information in the whole model. To qualitatively analyze the video-level representations, we project these last-layer features to the 2-dimensional space through t-SNE [25].

As shown in Figure 3a & 3a, in the cases without backgrounds (w/ BBox), our dynamic branch (the temporal stream) has better intra-class compactness and inter-class separability than the static one (the spatial stream). Though the dynamic branch shows overall superiority, the static stream is complementary to it: the dynamic branch sometimes confuses the orange data points with the yellow ones, while the static branch is capable of discriminating them correctly. In the case with backgrounds (w/o BBox), Figure 3c & 3d demonstrate that the background noises have negative impacts on both static and kinematic representations. In this case, our dynamic features are still highly robust to backgrounds compared with the static ones.

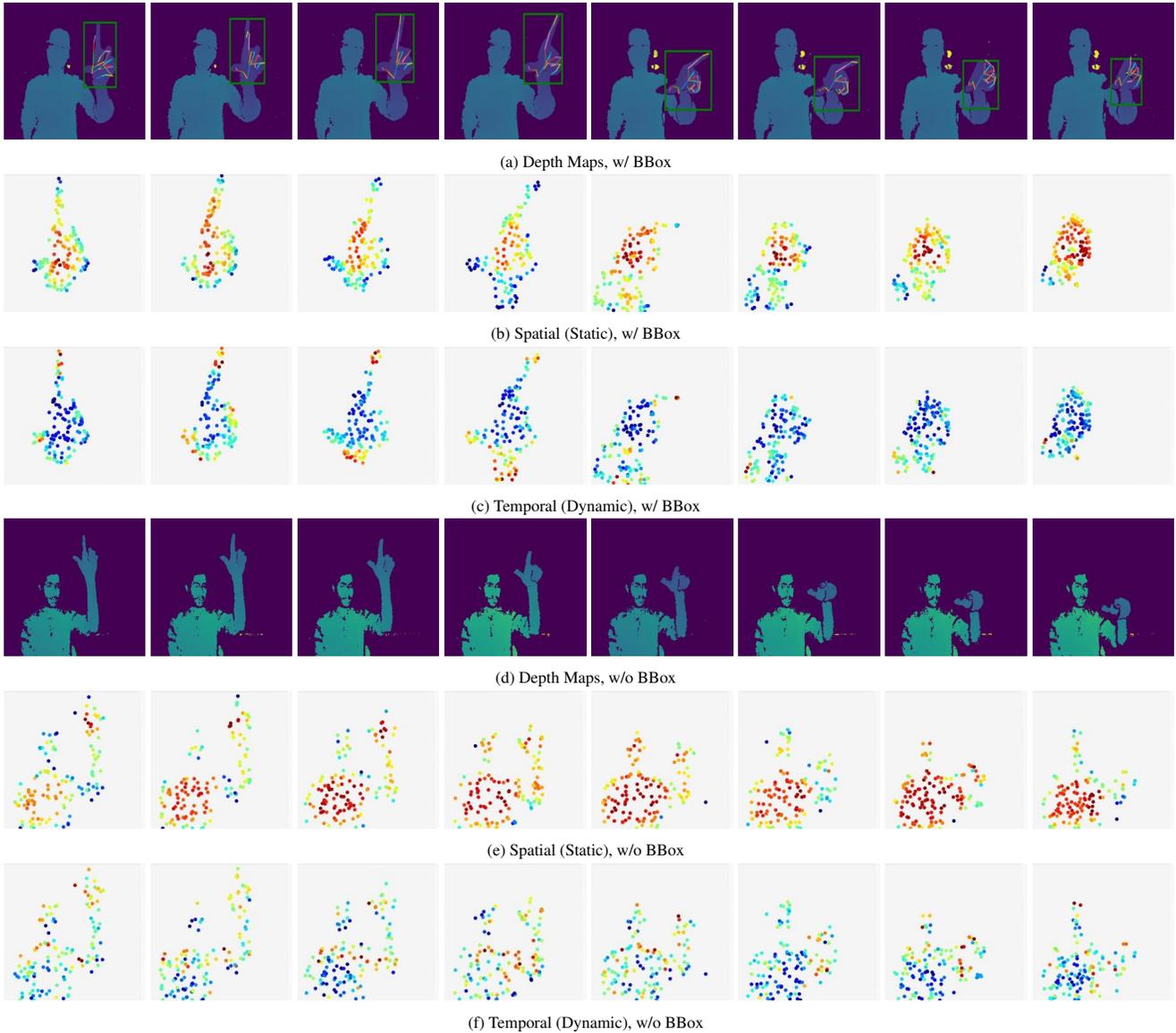

Figure 2. *Sequential raw depth inputs and point-level PACs on* SHREC'17. In PACs, the points in red have the highest activation values, while the blue ones are the lowest activating points. The animations of the above figures can be found in Figure 6 in the main body of our paper.

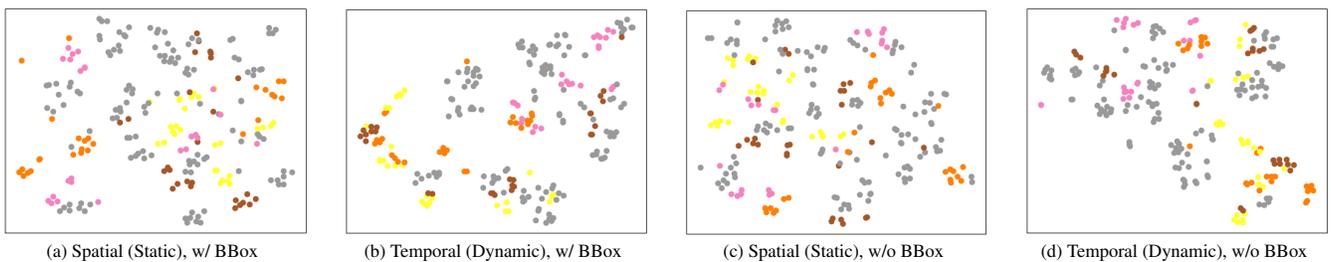

(a) Spatial (Static), w/ BBox  (b) Temporal (Dynamic), w/ BBox  (c) Spatial (Static), w/o BBox  (d) Temporal (Dynamic), w/o BBox

Figure 3. *Feature visualizations on* SHREC'17 *with video-level t-SNE*. For clarity, 8 out of the 28 classes are presented in the above figures. A video is visualized as a data point, of which the same color means the identical ground-truth category. The intra-class compactness and the inter-class separability reflect the representative ability of a model.

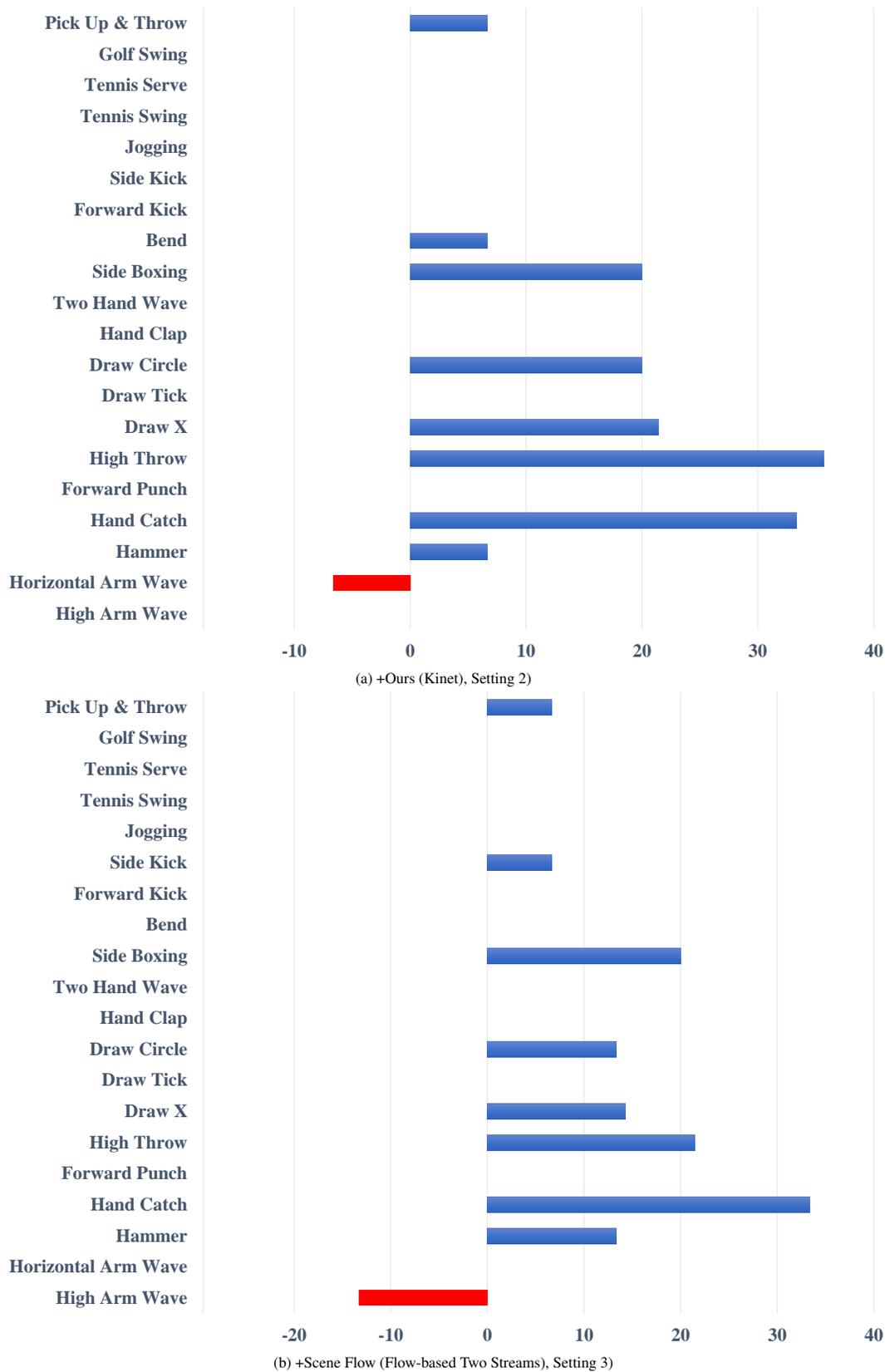

Figure 4. *Per-class accuracy gains (%) over the static model on 16-frame* MSRAction-3D. The blue bar indicates the postive change, whereas the red one represents the negative change.

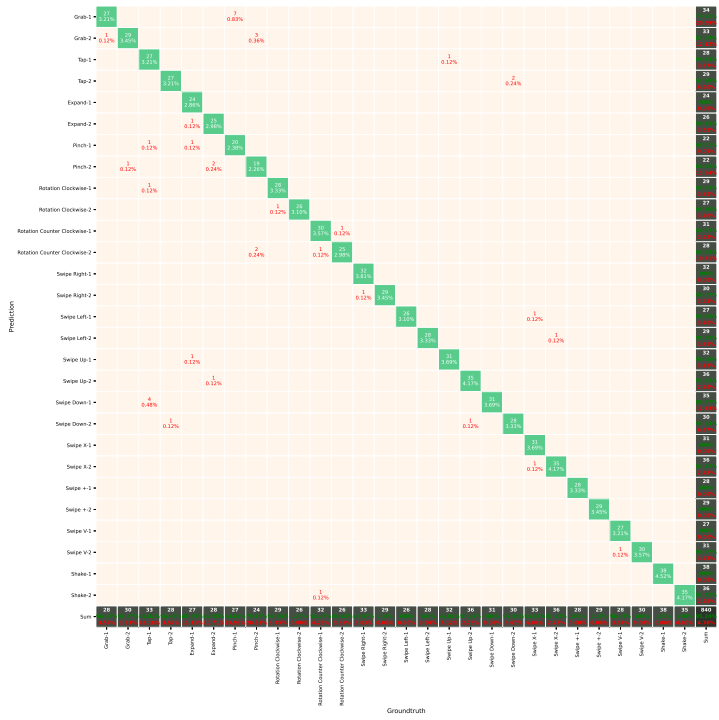

(a) w/ BBox

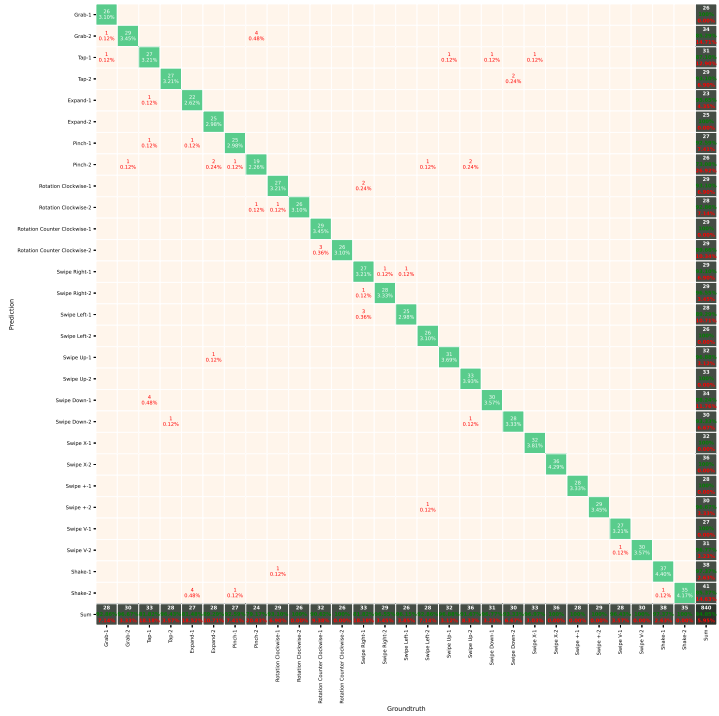

(b) w/o BBox

Figure 5. *Confusion matrices on* SHREC'17.

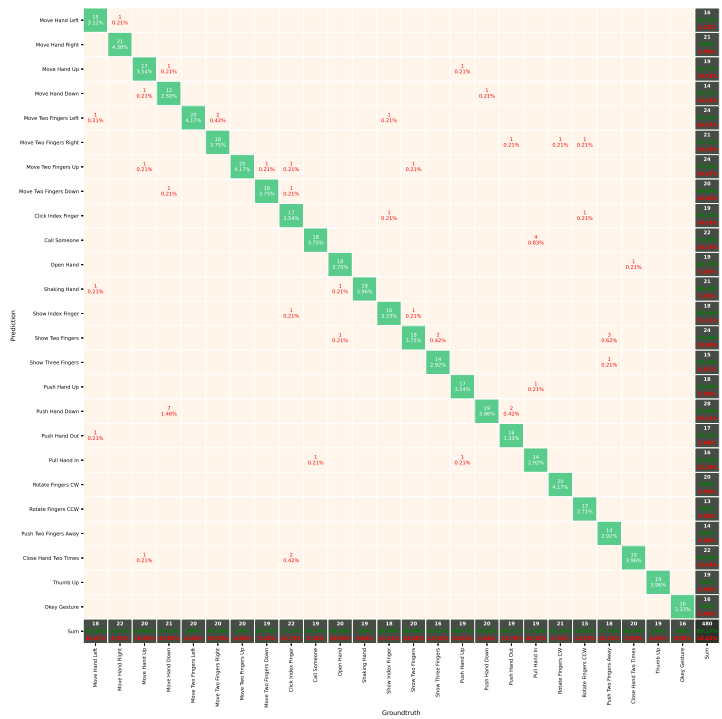

Figure 6. *Confusion matrix on* NvGesture.


\end{document}